\documentclass[hidelinks]{article}

\usepackage{arxiv}

\usepackage[utf8]{inputenc} 
\usepackage[T1]{fontenc}    

\usepackage{amsmath, amssymb} 

\usepackage{hyperref}       
\usepackage{url}            
\usepackage{amsfonts}       
\usepackage{nicefrac}       
\usepackage{microtype}      
\usepackage{graphicx}
\usepackage{doi}
\usepackage{natbib} 
\usepackage{pdflscape} 
\usepackage{acro} 
\usepackage{bookmark} 
\usepackage{nameref} 

\usepackage{tabularx}
\usepackage{booktabs}
\usepackage{caption}
\usepackage{tabu}
\usepackage{rotating} 
\usepackage{makecell} 
\usepackage{multirow} 
\usepackage{float} 
\usepackage{array}
\usepackage{arydshln}
\newcolumntype{P}[1]{>{\centering\arraybackslash}p{#1}}
\newcolumntype{R}[1]{>{\raggedleft\arraybackslash}p{#1}}

\usepackage{chngcntr}
\counterwithin{figure}{section}
\counterwithin{table}{section}
\counterwithin{equation}{section}

\usepackage[dvipsnames]{xcolor} 

\usepackage{algorithm}
\usepackage{algpseudocode}

\usepackage{bm} 

\usepackage{amsthm}

\title{Variational Quantum Circuit-Based Reinforcement Learning for Dynamic Portfolio Optimization}

\author{
    \href{https://orcid.org/0009-0002-1502-3670}
    {\includegraphics[scale=0.06]{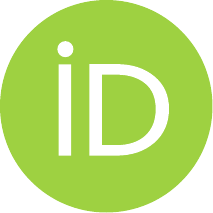}
    Vincent Gurgul} \\
	Chair of Information Systems\\
	Humboldt University Berlin\\
	Unter d. Linden 6, 10117 Berlin\\
	\And
	\href{https://orcid.org/0000-0002-2577-7348}
    {\includegraphics[scale=0.06]{orcid.pdf}
	Ying Chen} \\
	Department of Mathematics\\
	National University of Singapore\\
	2 Science Dr., Singapore 117543\\
	\And
	\href{https://orcid.org/0000-0001-7685-262X}
    {\includegraphics[scale=0.06]{orcid.pdf}
    Stefan Lessmann} \\
	Chair of Information Systems\\
	Humboldt University Berlin\\
	Unter d. Linden 6, 10117 Berlin\\
	Bucharest University of Economic Studies\\
    6 Piata Romana, 010374, Romania\\
}

\renewcommand{\headeright}{}
\renewcommand{\shorttitle}{Variational Quantum Circuit-Based Reinforcement Learning for Dynamic Portfolio Optimization}

\hypersetup{
pdftitle={Variational Quantum Circuit-Based Reinforcement Learning for Dynamic Portfolio Optimization},
pdfsubject={quantum, computing},
pdfauthor={Gurgul, V., Chen, Y., Lessmann, S.},
pdfkeywords={quantum finance, quantum machine learning, reinforcement learning, portfolio optimization},
}

\DeclareAcronym{MVO}{
	short = MVO,
	long = Mean-Variance Optimization
}
\DeclareAcronym{RL}{
	short = RL,
	long = Reinforcement Learning
}
\DeclareAcronym{VQC}{
	short = VQC,
	long = Variational Quantum Circuit
}
\DeclareAcronym{DDPG}{
	short = DDPG,
	long = Deep Deterministic Policy Gradient
}
\DeclareAcronym{QPU}{
	short = QPU,
	long = Quantum Processing Unit
}
\DeclareAcronym{PQC}{
	short = PQC,
	long = Parameterized Quantum Circuit
}
\DeclareAcronym{QNN}{
	short = QNN,
	long = Quantum Neural Network
}
\DeclareAcronym{QRL}{
	short = QRL,
	long = Quantum Reinforcement Learning
}
\DeclareAcronym{DQN}{
	short = DQN,
	long = Deep Q-Network
}
\DeclareAcronym{QUBO}{
	short = QUBO,
	long = Quadratic Unconstrained Binary Optimization
}
\DeclareAcronym{QAOA}{
	short = QAOA,
	long = Quantum Approximate Optimization Algorithm
}
\DeclareAcronym{VQE}{
	short = VQE,
	long = Variational Quantum Eigensolver
}
\DeclareAcronym{NISQ}{
	short = NISQ,
	long = Noisy Intermediate-Scale Quantum
}
\DeclareAcronym{CQM}{
	short = CQM,
	long = Constrained Quadratic Model
}
\DeclareAcronym{QA}{
	short = QA,
	long = Quantum Annealing
}
\DeclareAcronym{TD}{
	short = TD,
	long = Temporal-Difference
}
\DeclareAcronym{DP}{
	short = DP,
	long = Dynamic Programming
}
\DeclareAcronym{MDP}{
	short = MDP,
	long = Markov Decision Process
}

\begin{document}
\maketitle

\begin{abstract}

This paper presents a Quantum Reinforcement Learning (QRL) solution to the dynamic portfolio optimization problem based on Variational Quantum Circuits.
The implemented QRL approaches are quantum analogues of the classical neural-network-based Deep Deterministic Policy Gradient and Deep Q-Network algorithms.
Through an empirical evaluation on real-world financial data, we show that our quantum agents achieve risk-adjusted performance comparable to, and in some cases exceeding, that of classical Deep RL models with several orders of magnitude more parameters.
However, while quantum circuit execution is inherently fast at the hardware level, practical deployment on cloud-based quantum systems introduces substantial latency, making end-to-end runtime currently dominated by infrastructural overhead and limiting practical applicability.
Taken together, our results suggest that QRL is theoretically competitive with state-of-the-art classical reinforcement learning and may become practically advantageous as deployment overheads diminish.
This positions QRL as a promising paradigm for dynamic decision-making in complex, high-dimensional, and non-stationary environments such as financial markets.
The complete codebase is released as open source at: \url{https://github.com/VincentGurgul/qrl-dpo-public}

\end{abstract}



\section{Introduction}

Portfolio optimization is a cornerstone of finance, aiming to allocate assets in a way that maximizes returns while minimizing risk.
Traditional methods like \ac{MVO} have long been the standard approach for static portfolio allocation.
However, since such static methods treat portfolio optimization as a one-shot problem (an allocation is performed at single point in time based on past data), they often fail to capture the dynamic and sequential nature of real-world investment scenarios where market conditions are constantly evolving.

Recent advances in quantum computing offer potential for solving complex optimization problems more efficiently than classical counterparts.
Quantum Annealing and Variational Quantum Algorithms have shown promise in addressing static portfolio optimization formulations through the \ac{QUBO} framework.
While these quantum methods have demonstrated improvements over classical algorithms in solving QUBOs, the QUBO is also a static problem formulation and does not naturally enable adaptive decision-making over time.

\ac{RL}, a branch of machine learning that deals with sequential decision-making, has shown significant potential in dynamic portfolio management by facilitating the learning of optimal policies through interaction with the environment.
However, while powerful, \ac{RL} faces challenges in terms of scalability and training efficiency in the case of large state-action spaces.
\ac{QRL} merges \ac{RL} techniques with quantum computing, offering the potential for faster convergence and more efficient policy representation compared to classical methods.

Despite the growing interest in quantum technologies, there remains a notable gap in applying them to dynamic portfolio optimization.
Most existing studies in the quantum realm either treat portfolio problems as a static \ac{QUBO} and solve it using quantum solvers, or explore \ac{QRL} on simple control tasks without addressing complex real-world applications like sequential investment decisions.

This paper aims to bridge this gap by introducing a novel \ac{QRL} framework for dynamic portfolio optimization and systematically benchmarking it against classical baselines.
We propose a fully quantum \ac{RL} approach where the agent learns an optimal trading policy based on a hybrid optimization loop over a dynamic state-action space, while harnessing the entanglement-induced expressivity and potential for parallelism that is inherent to \acp{VQC}.

By integrating quantum computing into a sequential financial decision-making process, we demonstrate how this technology can enhance and accelerate traditional portfolio management practices and pave the way for more advanced investment strategies, but also highlight the current limitations of quantum hardware and cloud computing solutions.

\section{Literature Review}

In the following we contextualize our work within the landscape of portfolio optimization, \ac{RL}, and quantum machine learning by examining the evolution and current state of research across these intersecting fields.
We summarize foundational methods in classical portfolio theory, highlight recent advances in quantum algorithms for static optimization, and identify the emerging role of \ac{QRL}.
By identifying gaps in the exisiting literature we motivate our proposed QRL framework for portfolio optimization.

Portfolio optimization is a fundamental problem in finance, describing the process of allocating capital across a set of financial assets.
This objective is typically formulated as identifying the optimal portfolio weights \(\mathbf{w} = [w_1, \dots, w_N]\) over $N$ assets to balance expected return and risk. The majority of modern portfolio optimization methodologies build on the foundational model of \citet{markowitz_portfolio_1952}, in which the balance between expected return and portfolio risk is expressed as a constrained quadratic optimization problem:

\vspace{-12pt}
\begin{align}
\label{eq:markowitz_continuous}
	\max_{\mathbf{w}} \quad 
	& \boldsymbol{\mu}^\top \mathbf{w}
	- \eta\, \mathbf{w}^\top \boldsymbol{\Sigma}\,\mathbf{w}
	~\quad \text{s.t.} \quad \sum_{i=1}^N w_i = 1,~ w_i \ge 0,~ \eta \in [0,1] ~, \\[6pt]\nonumber
	\text{where} \quad 
	& \boldsymbol{\mu}^\top \mathbf{w} \text{ : portfolio return}, \\[2pt]\nonumber
	& \mathbf{w}^\top \boldsymbol{\Sigma}\,\mathbf{w} \text{ : portfolio variance (risk)}.
\end{align}
\vspace{-5pt}

In the Black--Litterman model, the Markowitz formulation is extended by replacing the return vector \(\boldsymbol{\mu}\) with a posterior estimate \(\boldsymbol{\mu}_{\text{BL}}\), which combines equilibrium market returns with subjective investor views \citep{black_global_1992}.
This approach stabilizes portfolio optimization by reducing the sensitivity to input return estimates while enabling the explicit incorporation of informed opinions.
The posterior mean of expected returns is given by

\vspace{-12pt}
\begin{align}
\label{eq:black_litterman}
	&\boldsymbol{\mu}_{\text{BL}} 
	= \boldsymbol{\pi} 
	+ \tau \boldsymbol{\Sigma} P^\top
	(P \tau \boldsymbol{\Sigma} P^\top + \Omega)^{-1}
	(\mathbf{q} - P\boldsymbol{\pi}), \\[15pt]\nonumber
	\text{where} \quad 
	& \boldsymbol{\pi} = \lambda \boldsymbol{\Sigma} \mathbf{w}_m : \text{ implied equilibrium returns,} \\[4pt]\nonumber
	& P : \text{ view matrix defining assets involved in each investor view,} \\[4pt]\nonumber
	& \mathbf{q} : \text{ vector of expected returns associated with investor views,} \\[4pt]\nonumber
	& \Omega : \text{ covariance matrix representing uncertainty of the views,} \\[4pt]\nonumber
	& \tau : \text{ scalar reflecting the uncertainty of the prior equilibrium returns.}
\end{align}
\vspace{-5pt}

For experienced investors, this framework allows the integration of domain knowledge and market intuition into quantitative allocation. However, in empirical or data-driven applications, a key challenge lies in quantitatively inferring coherent ``views'' from market sentiment or alternative data sources, introducing an additional layer of model complexity.

The Markowitz \ac{CQM} can also easily be extended to include additional constraints, such as limits on individual asset weights, sector exposures, transaction costs, $\text{CO}_2$ emissions, or ESG criteria \citep{fabozzi_robust_2012}. \cite{salirrosas_market_2025} extend the classic Markowitz formulation with binary asset weights by adding a market sentiment variable that increases diversification during ``anxious'' market sentiment periods, but find no advantage over the standard formulation.

Classical static optimization methods have a long history of being used to solve these problems \citep{salo_fifty_2024}.
However, dynamically rebalancing a portfolio over time (e.g.\ across many trading periods) is a complex decision-making task that challenges classical optimization techniques.
Recent advances in \ac{RL} offer powerful data-driven solutions for such dynamic portfolio management.
Deep neural network-based RL algorithms, particularly \acp{DQN}, have achieved state-of-the-art performance in portfolio trading and asset allocation, outperforming traditional strategies in terms of both Sharpe ratio and cumulative returns \citep{zhang_deep_2020}.

At the same time, quantum computing has emerged as a promising paradigm for tackling computationally hard optimization problems in finance. Leveraging superposition and entanglement, quantum algorithms can explore vast solution spaces and identify high-quality approximate solutions to NP-hard optimization problems, within practical runtime limits. While many of these approaches were initially described only theoretically, recent advances in \acp{QPU} have enabled the implementation on real hardware, albeit with limitations regarding noise and qubit counts \citep{orus_quantum_2019, egger_quantum_2020, buonaiuto_best_2023}.

\ac{QRL} lies at the intersection of these trends, aiming to integrate quantum computing into the RL framework. It is an emerging field with the potential to enhance learning and decision-making processes by using quantum circuits as function approximators or by accelerating parts of RL algorithms \citep{meyer_survey_2024}. In this section, we review prior work on quantum computing for portfolio optimization and on \ac{QRL}, followed by a synthesis of the key research gaps that motivate our study.

\subsection{Quantum Portfolio Optimization}

Recent reviews highlight three main strands of research in quantum portfolio optimization: the \ac{VQE}, the \ac{QAOA}, and \ac{QA} approaches, particularly those implemented on D-Wave hardware \citep{yulianti_implementation_2022, sehrawat_quantum_2024, volpe_improving_2025}.

A unified mathematical formulation underlies all three of these approaches: they require the optimization to be expressed as a \ac{QUBO} problem \citep{buonaiuto_best_2023}.
A QUBO is a quadratic objective function defined over binary decision variables without explicit constraints:

\vspace{-10pt}
\begin{align*}
	\min_{\mathbf{b}} \quad 
	& \mathbf{b}^\top \mathbf{Q}\, \mathbf{b} + \mathbf{c}^\top \mathbf{b}~, \\[6pt]
	\text{where} \quad 
	& \mathbf{b} \in \{0,1\}^N \text{ : binary decision variables}, \\[4pt]
	& \mathbf{Q} \in \mathbb{R}^{N \times N} \text{ : symmetric matrix of quadratic coefficients}, \\[4pt]
	& \mathbf{c} \in \mathbb{R}^N \text{ : vector of linear coefficients}.
\end{align*}

Despite its apparent simplicity, the \ac{QUBO} formalism is remarkably general.
Any problem defined over a finite set of discrete variables, including all NP-hard combinatorial optimization problems, can be reformulated exactly in this structure.
Continuous or mixed-integer problems can be approximated through binary encodings or quadratic penalty functions that emulate continuous behaviour \citep{kochenberger_unconstrained_2014, glover_tutorial_2019}.

Portfolio optimization can be formulated in QUBO form through two approaches, both static in nature.
The first treats portfolio optimization as a combinatorial selection problem, where the goal is to select a subset of assets from a larger pool to maximize return and minimize risk.
This involves defining binary decision variables $b_i$ that indicate whether asset $i$ is included in the portfolio ($b_i = 1$) or not ($b_i = 0$).
Alternatively, it can be viewed as going long or short on each asset, with $b_i = 1$ indicating a long position and $b_i = 0$ indicating a short position.
The objective function can then be written as a constrained quadratic function of these binary variables, as expressed in Eq.~\ref{eq:binary_markowitz}, incorporating expected returns and covariances between asset returns \citep{phillipson_portfolio_2021, brandhofer_benchmarking_2022}.

\vspace{-10pt}
\begin{align}
\label{eq:binary_markowitz}
	\max_{\mathbf{b}} \quad 
	& \boldsymbol{\mu}^\top \mathbf{b} - \eta\, \mathbf{b}^\top \boldsymbol{\Sigma}\,\mathbf{b} 
	~\quad \text{s.t.} \quad \sum_{i=1}^N b_i = K~, \\[4pt]\nonumber
	\text{where} \quad 
	& \mathbf{b} \in \{0,1\}^N \text{ : binary asset selection vector}, \\[2pt]\nonumber
	& b_i = 1 \text{ if asset $i$ is included in the portfolio, else } 0, \\[2pt]\nonumber
	& \eta \in \mathbb{R} \text{ : risk aversion parameter}, \\[2pt]\nonumber
	& \boldsymbol{\mu}^\top \mathbf{b} \text{ : expected return of selected assets}, \\[2pt]\nonumber
	& \mathbf{b}^\top \boldsymbol{\Sigma}\,\mathbf{b} \text{ : variance (risk) of the selected portfolio}, \\[2pt]\nonumber
	& K \text{ : predefined number of assets to be selected.}
\end{align}

Since quantum algorithms such as the \ac{VQE}, \ac{QAOA}, and \ac{QA} operate on unconstrained objective functions, the cardinality constraint must be incorporated as a quadratic penalty term, yielding the unconstrained (QUBO) formulation in Eq.~\ref{eq:binary_portfolio_qubo}.

\vspace{-8pt}
\begin{equation}
\label{eq:binary_portfolio_qubo}
	\max_{\mathbf{b}} \;
	\boldsymbol{\mu}^\top \mathbf{b}
	- \eta\, \mathbf{b}^\top \boldsymbol{\Sigma}\,\mathbf{b}
	- \lambda \left(\sum_{i=1}^N b_i - K\right)^2\,,
\end{equation}

where $\lambda \in \mathbb{R}_{>0}$ is a penalty coefficient ensuring approximate satisfaction of the constraint.

The second approach treats portfolio optimization as a multi-objective problem of allocating fractions of a portfolio value -- expressing the portfolio problem as a \ac{CQM} (the classic Markowitz formulation, see Eq.~\ref{eq:markowitz_continuous}).

This formulation can be discretized by replacing continuous asset weights $w_i$ with the closest integer asset amount $n_i$ corresponding to the price of the asset divided by the amount of capital to be invested.
Hence, the following binarization approach applies also to discrete Markowitz formulations.

With the help of a binary conversion matrix $\mathbf{C}$, the discrete/discretized integer portfolio optimization can be encoded into binary decision variables, allowing the problem to be represented as a binary quadratic optimization problem.
Finally, to arrive at \ac{QUBO} form, the constraints need to be included directly in the optimization as a penalty term \citep{buonaiuto_best_2023}.
The resulting formulation is expressed in Eq.~\ref{eq:binarized_portfolio_qubo}.

\begin{align}
\label{eq:binarized_portfolio_qubo}
\max_{\mathbf{b}} \quad 
& (\mathbf{p^*}\circ\boldsymbol{\mu})^\top \mathbf{C}\, \mathbf{b}
- \eta\, \mathbf{b}^\top \mathbf{C}^\top \big( (\mathbf{p^*}\circ\boldsymbol{\Sigma})^\top\circ\mathbf{p^*} \big)\, \mathbf{C}\, \mathbf{b} - \lambda((\mathbf{C}^\top\mathbf{p^*})^\top\,\mathbf{b}-1)~, \\[6pt]\nonumber
\text{where} \quad
& \mathbf{b} \in \{0,1\}^M \,,~ M = \sum_{i=1}^N d_i \text{ : binarized target vector}\,, \\[6pt]\nonumber
& \lambda \in \mathbb{R}_{>0} \text{ : penalty coefficient for budget constraint}\,, \\[6pt]\nonumber
& d_i = \mathit{Int}(\log_2(n_i^{\max})) \text{~~where~~} \mathbf{n} : \text{ integer asset amounts}\,, \\[8pt]\nonumber
& \mathbf{C} =
\begin{pmatrix}
2^0 & \cdots & 2^{d_1} & 0 & \cdots & 0 \\[2pt]\nonumber
0 & \cdots & 0 & 2^0 & \cdots & 2^{d_2} \\[2pt]\nonumber
\vdots & \ddots & \vdots & \vdots & \ddots & \vdots \\[2pt]\nonumber
0 & \cdots & 0 & 0 & \cdots & 2^{d_N}
\end{pmatrix}\,,\\[6pt]\nonumber
& \mathbf{p^*} =  \frac{\mathbf{p}}{B} \text{ : relative asset prices} \\[6pt]\nonumber
& \mathbf{p} \in \mathbb{R}^N \text{ : asset prices}\,, \quad B \in \mathbb{R} \text{ : portfolio budget}\,.\nonumber
\end{align}

Once expressed in any of the mentioned \ac{QUBO} forms, the objective can be converted into a set of quantum operators (Ising Hamiltonian) and subsequently solved using different quantum paradigms \citep{egger_quantum_2020}, such as \ac{QAOA}, \ac{VQE}, and \ac{QA}. An overview of existing literature on quantum portfolio optimization, including problem formulations, approaches, and implementation details, is provided in Table~\ref{tab:quantum_portfolio_lit}.

The \ac{QAOA} alternates between two parameterized quantum operators: a cost Hamiltonian encoding the objective function and a mixer Hamiltonian that explores the feasible space \citep{farhi_quantum_2014}. By iteratively adjusting the parameters of these unitaries through a classical optimization loop, \ac{QAOA} can approximate a low-energy configuration that corresponds to the optimal portfolio allocation, although the restrictive assumptions under which such behaviour could be formally established have not been validated \citep{fakhimi_quantum_2023}. Despite these theoretical limitations, \ac{QAOA} has been successfully implemented on quantum hardware to solve small-scale portfolio optimization problems \citep{stopfer_quantum_2025, herman_constrained_2023}.

The \ac{VQE} follows a similar hybrid quantum-classical paradigm but leverages the variational principle to estimate the minimum eigenvalue of the Ising Hamiltonian. In this framework, a \ac{PQC} prepares a trial wavefunction whose expected energy, computed as the expectation value of the Hamiltonian, is minimized through classical optimization. The \ac{VQE} algorithm thus yields a solution corresponding to the minimum energy, or optimal portfolio configuration \citep{peruzzo_variational_2014}. Recent implementations on IBM's superconducting quantum hardware demonstrate the feasibility of this approach on small-scale portfolio instances \citep{wang_achieving_2025, herman_constrained_2023}. \cite{kolotouros_evolving_2022} implement both \ac{QAOA} and \ac{VQE} for portfolio optimization on simulators, proposing an evolving objective function to improve convergence and nicely visualizing optimization landscapes. \cite{herman_constrained_2023} compare \ac{QAOA} and \ac{VQE} for constrained portfolio optimization, proposing an approach to include multiple contraints into the problem formulation. \cite{chen_black-litterman_2023} integrate the Black-Litterman model into the quantum domain by using a \ac{VQC} for modelling investor views, combined with \ac{VQE} for addressing the asset selection problem. \cite{zaman_po-qa_2024} systematically benchmark the impact of hyperparameters such as rotation blocks, repetitions, and entanglement types on the performance of \ac{QAOA} and \acp{VQE} for portfolio optimization.

\ac{QA}, by contrast, realizes the optimization directly on hardware through the adiabatic evolution of a quantum system from an easily prepared ground state toward the ground state of the Ising Hamiltonian. Similarly to the \ac{VQE} approach, the system's lowest-energy state corresponds to the optimal asset allocation \citep{kadowaki_quantum_1998, das_colloquium_2008}. \cite{tang_comparative_2024} compare 14 \ac{QA} and reverse \ac{QA} approaches, finding no significant difference in time-to-solution, but an improved success probability with coupler and counter-diabatic \ac{QA}. Further early demonstrations on D-Wave devices have demonstrate that quantum annealers can efficiently explore large combinatorial search spaces and produce competitive approximate solutions, especially when integrated with classical pre- and post-processing routines \citep{rosenberg_solving_2016, venturelli_reverse_2019,cohen_portfolio_2020}.

\begin{landscape}
\begin{table}
    \centering
    \footnotesize
    \caption{Overview of quantum portfolio optimization literature (chronological order)}
    \vspace{0.8em}
    \renewcommand{\arraystretch}{1.5}
    \begin{tabular}{|l|l|l|l|c|c|c|}
    \hline
    \textbf{Paper} &
    \textbf{Problem Formulation} &
    \textbf{Approach} &
    \textbf{Implementation} &
	\rotatebox{90}{\parbox{2.2cm}{\textbf{Classical\\Benchmarks}}} &
    \rotatebox{90}{\textbf{Dynamic}} &
    \rotatebox{90}{\parbox{2.85cm}{\textbf{Forward-looking~}}} \\
    \hline
    \cite{rosenberg_solving_2016} & Static discrete Markowitz (see Eq. \ref{eq:binarized_portfolio_qubo}) & QA & QPU & & & \\
    \hline
    \cite{venturelli_reverse_2019} & Static binary Markowitz (see Eq. \ref{eq:binary_portfolio_qubo}) & QA & QPU & \checkmark & & \\
    \hline
    \cite{cohen_portfolio_2020} & Static binary Markowitz (see Eq. \ref{eq:binary_portfolio_qubo}) & QA & QPU & \checkmark & & \\
    \hline
    \cite{phillipson_portfolio_2021} & Static binary Markowitz (see Eq. \ref{eq:binary_portfolio_qubo}) & QA & QPU & \checkmark & & \\
    \hline
    \cite{mugel_dynamic_2022} & Static binary Markowitz (see Eq. \ref{eq:binary_portfolio_qubo}) & QA and VQE & QPU & \checkmark & \checkmark & \\
    \hline
    \cite{kolotouros_evolving_2022} & Static binary Markowitz (see Eq. \ref{eq:binary_portfolio_qubo}) & QAOA and VQE & Simulator & & & \\
    \hline
    \cite{brandhofer_benchmarking_2022} & Static binary Markowitz (see Eq. \ref{eq:binary_portfolio_qubo}) & QAOA & QPU & & & \\
    \hline
    \cite{heras_backtesting_2023} & Static binary Markowitz (see Eq. \ref{eq:binary_portfolio_qubo}) & VQE & QPU & \checkmark &  &\\
    \hline
    \cite{herman_constrained_2023} & Static continuous Markowitz (see Eq. \ref{eq:binarized_portfolio_qubo}) & QAOA and VQE & Simulator and QPU & &  & \\
    \hline
    \cite{jain_efficient_2023} & Static binary Markowitz (see Eq. \ref{eq:binary_portfolio_qubo}) & Hybrid Annealer-Gate & Simulator & & & \\
    \hline
    \cite{buonaiuto_best_2023} & Static continuous Markowitz (see Eq. \ref{eq:binarized_portfolio_qubo}) & VQE & QPU & \checkmark & & \\
    \hline
    \cite{chen_black-litterman_2023} & Static binary Black-Litterman (see Eqs. \ref{eq:binary_portfolio_qubo} \& \ref{eq:black_litterman})& VQC + VQE & Simulator & \checkmark & & \\
    \hline
    \cite{aguilera_multi-objective_2024} & Static discrete Markowitz + $\text{CO}_2$ emission constraints & QA & Simulator and QPU & \checkmark & & \\
    \hline
    \cite{catalano_quantum_2024} & Static discrete Markowitz + ESG constraints & QA & QPU & & & \\
    \hline
    \cite{quynh_quantum_2024} & Static binary Markowitz (see Eq. \ref{eq:binary_portfolio_qubo}) & Extension of VQE/QAOA & Simulator & & & \\
    \hline
    \cite{tang_comparative_2024} & Static binary Markowitz (see Eq. \ref{eq:binary_portfolio_qubo}) & Various variants of QA & Simulator & \checkmark & & \\
    \hline
    \cite{chou_exploring_2024} & Static discrete Markowitz (see Eq. \ref{eq:binarized_portfolio_qubo}) & QA & QPU & \checkmark & & \\
    \hline
    \cite{zaman_po-qa_2024} & Static binary Markowitz (see Eq. \ref{eq:binary_portfolio_qubo}) & QAOA and VQE & Simulator & \checkmark & & \\
    \hline
    \cite{salirrosas_market_2025} & Static binary Markowitz + market sentiment indicator & VQE & Simulator & & & \\
    \hline
    \cite{wang_variational_2025} & Static binary Markowitz (see Eq. \ref{eq:binary_portfolio_qubo}) & VQE & Simulator and QPU & & & \\
    \hline
    \cite{wang_achieving_2025} & Static continuous Markowitz (see Eq. \ref{eq:binarized_portfolio_qubo}) & QA & QPU & \checkmark & & \\
    \hline
    \cite{stopfer_quantum_2025} & Static binary Markowitz (see Eq. \ref{eq:binary_portfolio_qubo}) & QA and QAOA & QPU & \checkmark & & \\
    \hline
    The proposed approach & Dynamic continuous state-action space (see Eq. \ref{eq:state_action_space}) & VQC-RL & Simulator + QPU & \checkmark & \checkmark & \checkmark \\
    \hline
    \end{tabular}
	\\\vspace{7pt}
	\raggedright\scriptsize{Note: In the context of this table, ``and'' indicates that multiple approaches were implemented and evaluated side by side, whereas ``+'' signifies that the components were used in conjunction as part of a single approach.}
    \label{tab:quantum_portfolio_lit}
\end{table}
\end{landscape}


However, practical performance is still shown to be limited by noise, qubit connectivity, and embedding constraints \citep{phillipson_portfolio_2021, sakuler_real-world_2025}. A recent benchmark study by \cite{stopfer_quantum_2025} sheds further doubt on the practical utility of \ac{QA} and \ac{QAOA} for portfolio optimization, showing that classical heuristics can solve small portfolio optimization problems in seconds while outperforming quantum approaches in solution quality, indicating that current hardware does not yet facilitate a quantum advantage in solving complex QUBO formulations.

In addition, \cite{stopfer_quantum_2025} show in their benchmarking study that modern classical solvers can still exactly solve small portfolio optimization problems (tens of assets) in seconds, and a tailored classical heuristic outperforms quantum approaches in solution quality, indicating that current hardware does not yet facilitate a quantum advantage in portfolio optimization.

It's also important to note, that, similarly to classical \ac{MVO}, all of the mentioned approaches are confined to static optimization formulations (solving for optimal asset allocation for a given time point). A notable effort tackling dynamic or sequential portfolio decisions using quantum techniques is \cite{mugel_dynamic_2022}, who implemented several quantum and quantum-inspired algorithms for dynamic portfolio optimization on real-world data. They tested D-Wave's hybrid \ac{QA}, two VQE-based methods on IBM quantum processors, and a quantum-inspired tensor network method. This study was the first to apply quantum algorithms to a multi-step portfolio rebalancing scenario, highlighting both the potential and the challenges of moving beyond static use-cases. However, even in this dynamic study, the quantum approaches essentially solve a static encoding of the multi-period problem (by embedding multiple time steps into one large optimization problem) rather than learning an adaptive trading policy.

In summary, while there has been steady progress in applying quantum annealers and variational quantum algorithms to portfolio optimization, these efforts have so far focused on static optimization or multi-period static formulations, and classical methods still often hold an edge in performance under practical conditions. This leaves open the question of how to leverage quantum computing for truly adaptive, sequential decision-making in finance and whether a quantum advantage can be realized under those conditions.

\subsection{Quantum Reinforcement Learning}

\ac{QRL} is a nascent but growing area of research that seeks to merge RL techniques with quantum computing. In broad terms, \ac{QRL} algorithms use quantum resources to either accelerate RL computations or to represent policy/value functions in new ways \citep{meyer_survey_2024}. The most actively studied branch of QRL involves \acp{VQC} serving as function approximators (replacing classical counterparts in a RL algorithm). In these approaches, the agent interacts with a classical environment as usual, but the policy or Q-value function is encoded by a quantum circuit whose parameters are trained via classical optimization (loss backpropagation). This paradigm fits the capabilities of today's noisy intermediate-scale quantum (NISQ) devices and its feasibility has been demonstrated in standard control environments. For example, hybrid agents using a \ac{VQC} within a \ac{DQN} were shown to learn optimal actions in discrete tasks like FrozenLake\footnote{\textbf{FrozenLake:} A benchmark environment from OpenAI Gym that tests discrete-state, discrete-action control. The agent must navigate a grid world (a frozen lake) from a start to a goal location while avoiding holes.}, using significantly fewer parameters than an equivalent classical \ac{DQN} \citep{chen_variational_2020}. Similarly, \cite{lockwood_reinforcement_2020} incorporate \ac{VQC} approximators into value-based frameworks and demonstrate its ability to solve simple continuous control tasks like CartPole\footnote{\textbf{CartPole:} A control task where the agent must balance a pivoted pole attached to a controllable cart (left/right). The system provides continuous state observations (cart position, velocity, pole angle, and angular velocity).} and Blackjack, especially when efficient quantum state encoding schemes were employed. These results suggest that quantum models can represent policies or value functions with a compact encoding, offering sample-efficiency and faster convergence in some cases.

Beyond value-based \ac{RL}, prior work has also explored quantum variants of policy-gradient and Actor-Critic methods. \cite{jerbi_parametrized_2021} introduce quantum policy iteration and provide theoretical evidence for a quantum advantage in reinforcement learning for specific cases. \cite{lin_quantum_2025} integrate \acp{VQC} with non-local observables into \ac{DQN} and Asynchronous Advantage Actor-Critic frameworks. They show that a \ac{VQC}-enhanced agent can match or exceed the performance of a classical agent, with indications of faster learning thanks to the expressive power of quantum circuits. These works illustrate that QRL algorithms can be realized on current hardware and they hint at possible advantages, for example, using fewer parameters or achieving better sample efficiency. However, \ac{QRL} is still in an exploratory stage---most demonstrations so far are on small-scale or toy problems, and practical evidence of a quantum advantage in complex, real-world \ac{RL} tasks remains elusive. A notable application in finance is \cite{cherrat_quantum_2023} who apply \ac{QRL} to solve hedging problems. They implement quantum policy-search and distributional Actor-Critic algorithms for optimal hedging of financial derivatives (a sequential decision problem), using \acp{VQC} in the agent's policy and value functions. Their results show that \ac{QRL} could achieve performance comparable to classical Deep \ac{RL} while using significantly fewer trainable parameters, and the quantum model even outperformed classical baselines in certain metrics. Moreover, they successfully ran their QRL agent on a trapped-ion processor with 16 qubits, demonstrating the feasibility of QRL on present-day quantum hardware. Another finance-related example is the work of \cite{yang_apply_2023}, who explore a QRL approach for American option pricing, a complex sequential optimization problem without an analytical closed-form solution. By integrating quantum computing into a Deep RL framework for option pricing, they improve the efficiency of learning optimal exercise policies. While these studies underscore the potential of QRL in financial decision problems, portfolio optimization has not yet been addressed with \ac{QRL}, despite classical RL methods achieving state-of-the-art performance in this domain \citep{zhang_deep_2020}.

\subsection{Identified Gaps}

There is a lack of frameworks where a quantum or hybrid quantum-classical algorithm learns a policy over time in response to market state changes. In other words, the sequential nature of portfolio rebalancing has been largely absent from quantum approaches. The focus in the quantum domain so far has been on portfolio optimization formulated as a static \ac{QUBO}, leaving a void in addressing forward-looking dynamic decision-making. 

The majority of \ac{QRL} research has focused on basic control tasks, indicating a need for more development and practical demonstrations of \ac{QRL} algorithms on complex real-world problems. Existing literature lacks any demonstration of \ac{QRL} for portfolio optimization, despite the clear success of RL in classical portfolio management and the active interest in quantum methods for static portfolio optimization. The only finance-related QRL works so far have addressed option pricing and hedging.

In this paper, we aim to bridge these gaps by proposing a \ac{QRL} framework for portfolio optimization and demonstrating its implementation with real-world financial data on quantum hardware.


\section{Methodology}

\subsection{Foundations of Quantum Computing}

Quantum computing builds on the principles of quantum mechanics to provide a fundamentally different model of information processing compared to classical digital computation \citep{nielsen_quantum_2010}.
Whereas classical computers operate on bits that take values in $\{0,1\}$, quantum computers manipulate quantum bits, or \emph{qubits}, which are realized physically as controllable two-level quantum systems. 
Depending on the hardware platform, qubits can be encoded in electron spins, photon polarizations, trapped ions, or current loops in superconducting circuits. 
All of these implementations share the same mathematical structure: a state described by a vector in a two-dimensional complex Hilbert space.

A single qubit is most generally written as a coherent superposition of the computational basis states $|0\rangle$ and $|1\rangle$,
\begin{equation*}
    \boldsymbol{\psi} :~ \mathbb{C}^2 \rightarrow \mathcal{H}_2\,,
    \quad
    (\alpha,\beta) \,\mapsto\, \alpha|0\rangle + \beta|1\rangle,
    \quad \text{s.t.} \quad |\alpha|^2 + |\beta|^2 = 1\,.
\end{equation*}

where $\{|0\rangle, |1\rangle\}$ forms an orthonormal basis of the two-dimensional Hilbert space $\mathcal{H}_2$ and the coefficients $\alpha,\beta \in \mathbb{C}$ are called the probability amplitudes.
The normalization condition ensures that a measurement of the qubit in the computational basis yields either $|0\rangle$ or $|1\rangle$ with probabilities $|\alpha|^2$ and $|\beta|^2$, respectively. 
While this superficially resembles a probabilistic mixture, the state is fundamentally more expressive: the relative phase between $\alpha$ and $\beta$ enables constructive and destructive interference under unitary transformations of the qubit, a phenomenon with no classical analogue.

\begin{figure}[H]
	\centering
	\includegraphics[width=0.4\textwidth]{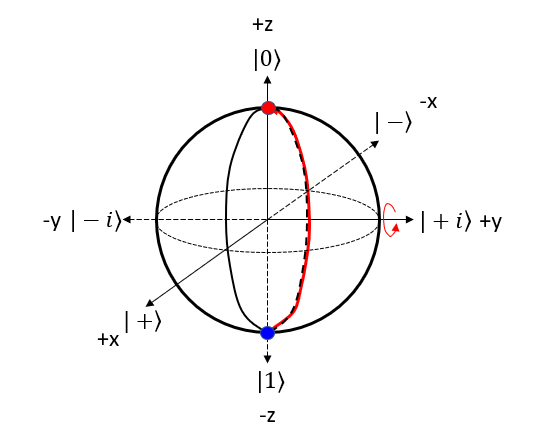}
	\caption{Visualization of the Bloch sphere \citep{freiman_hydrogen_2024}.}
	\label{fig:bloch}
\end{figure}

The geometry of qubits is most clearly seen through the Bloch sphere representation. 
Any pure qubit state can be re-parameterized as
\begin{equation*}
    \boldsymbol{\psi} 
    = \cos\!\left(\tfrac{\theta}{2}\right)\,|0\rangle
    + e^{i\phi}\sin\!\left(\tfrac{\theta}{2}\right)\,|1\rangle\,.
\end{equation*}

with angles $\theta \in [0,\pi]$ and $\phi \in [0,2\pi)$. 
This maps each qubit state to a unique point on the surface of a unit sphere, called the Bloch sphere. 
The computational basis states $|0\rangle$ and $|1\rangle$ occupy the north and south poles, respectively, while any other point on the sphere represents a coherent superposition $\alpha|0\rangle + \beta|1\rangle$ with relative phase and amplitude determined by its position (see Figure \ref{fig:bloch}).

\vspace{-5pt}

From a geometric perspective, quantum gates on a single qubit are rotations of this vector on the Bloch sphere. 
The abstract rotation operator
\begin{equation*}
    \mathbf{R}_{\hat{n}}(\theta) 
    = e^{-i \, \tfrac{\theta}{2} \, \hat{n} \boldsymbol{\sigma}}\,,
\end{equation*}
 
where $\boldsymbol{\sigma} = (\mathbf{X},\mathbf{Y},\mathbf{Z})$ denotes the Pauli matrices, represents a physical rotation of the qubit state by an angle $\theta$ about the Bloch-sphere axis $\hat{n}$, implemented physically through controlled electromagnetic pulses.

When multiple qubits are combined, the state space is given by the tensor product of the individual qubit spaces. 
A system of $n$ qubits is described by a vector in a $2^n$-dimensional Hilbert space. 
For example, a two-qubit state has the general form

\begin{equation*}
    \boldsymbol{\psi} =
    \alpha_{00}|00\rangle + \alpha_{01}|01\rangle 
    + \alpha_{10}|10\rangle + \alpha_{11}|11\rangle\,
    \quad \text{s.t.}\quad \sum |\alpha_{ij}|^2 = 1\,,
    \quad \boldsymbol{\psi} \in \mathcal{H}_4\,,
\end{equation*}

The distinguishing feature of such multi-qubit states is not merely their dimensionality, but the presence of complex phases across the amplitudes $\alpha_{ij}$. These phases give rise to coherent interference when the system is acted upon by unitary transformations. As the number of qubits increases, interference can occur across an exponentially growing set of basis states. This rapidly expanding interference structure is the fundamental source of the expressive power of multi-qubit quantum systems.

However, coherent interference across multiple qubits requires physical interactions between them.
In practice, such interactions are realized through two-qubit gates---most notably the Controlled-NOT (CNOT) gate, which conditionally flips the state of the second qubit (the target qubit) if and only if the first qubit (the control qubit) is $|1\rangle$. It thereby creates correlations between qubits that cannot be produced by single-qubit rotations alone.
Such correlations between qubits are referred to as entanglement and the gates that facilitate them are called entagling gates \citep{horodecki_quantum_2009}.

Quantum gates are designed to be universal, meaning that any unitary transformation on $n$ qubits can be decomposed into a sequence of single-qubit rotations and a fixed entangling gate such as CNOT \citep{barenco_elementary_1995}. 
On this foundation, several quantum algorithms have been developed that provably outperform their best-known classical counterparts. 
Examples include Shor’s algorithm for factoring large integers in polynomial time \citep{shor_polynomial-time_1997}, Grover’s algorithm for unstructured search offering quadratic speedup \citep{grover_quantum_1997}, and quantum simulation algorithms for chemistry and materials science \citep{aspuru-guzik_simulated_2005}. 
While portfolio optimization is not yet among the domains with a proven theoretical quantum advantage, the same principles of state space expressivity and interference apply to optimization and learning tasks.

However, realizing quantum computation in practice comes with multiple challenges.
Interactions between qubits and their environment lead to the loss of quantum phase information, a phenomenon referred to as decoherence.
Qubits must exhibit coherence times long enough to support computationally meaningful operations, while simultaneously remaining accessible to precise external control through electromagnetic pulses. 
Balancing isolation from the environment with controllability constitutes a central engineering challenge. 
In addition, all existing quantum devices are subject to noise: unwanted environmental interactions, gate errors caused by control pulses, and measurement errors. 
This defines the so-called \ac{NISQ} era, enabling machines with tens to hundreds of non-error-corrected qubits \citep{preskill_quantum_2018}. 
In this regime, fault-tolerant computation is not yet feasible, and algorithm design must explicitly account for noise.

\subsection{Variational Quantum Circuits and Quantum Neural Networks}

\acp{PQC} form the cornerstone of modern quantum machine learning and variational quantum algorithms.
A \ac{PQC} is a quantum circuit with tunable parameters in which a parameterized unitary transformation $\mathbf{U}(\boldsymbol{\theta})$ acts on an input state.
\acp{VQC} are a subclass of \acp{PQC} in which the tunable parameters are additionally optimized according to a cost function defined on measurement outcomes \citep{benedetti_parameterized_2019}.
This optimization is typically performed using gradient-based methods, where gradients can be estimated using techniques like the parameter-shift rule \citep{schuld_evaluating_2019}.
When viewed through the lens of machine learning, the parametrization and optimization of these circuits motivates the terminology \ac{QNN} \citep{cerezo_variational_2021}.
However, there are different approaches frequently lumped together under the term \ac{QNN}, encompassing models as diverse as quantum associative memories, quantum perceptrons, and interacting quantum-dot systems \citep{schuld_quest_2014}. 
We therefore avoid this ambiguous terminology and will henceforth refer to \acp{VQC}.

\vspace{-5pt}
\begin{figure}[H]
	\centering
	\includegraphics[width=0.61\textwidth]{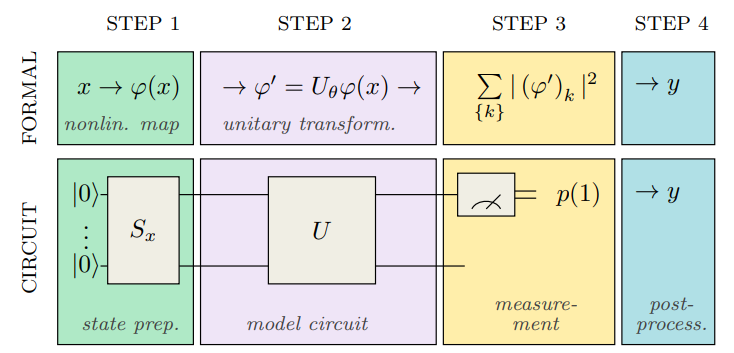}
	\caption{Illustration of the four steps involved in inference with a \ac{VQC} \citep[Adapted from][]{schuld_circuit-centric_2020-1}.}
	\label{fig:vqc}
\end{figure}

As illustrated in Fig.~\ref{fig:vqc}, the general workflow of a VQC consists of four fundamental steps: (i) \emph{state preparation}, where classical input data $\mathbf{x}$ is encoded into a quantum state, (ii) \emph{model transformation}, where the parametrized unitary $\mathbf{U}(\boldsymbol{\theta})$ acts on the encoded state, (iii) \emph{measurement}, where expectation values of observables are sampled from the output state, and (iv) \emph{post-processing}, where the measurement outcomes are, for example, used to compute a cost function.

In the first step, classical data must be embedded onto a quantum state. There are multiple strategies for this data encoding, and the choice of method strongly influences both the expressive capacity of the circuit and its resource requirements. Three common approaches are basis encoding, angle encoding, and amplitude encoding \citep{schuld_effect_2021}.

In \textbf{basis encoding}, the simplest type of classical data representation, each component of a binary string $\mathbf{x} \in \{0,1\}^n$ is directly mapped to one of the (discrete) computational basis states of an $n$-qubit register:

\vspace{-5pt}
\begin{equation*}
	\{0,1\}^n \to \mathcal{H}_{2^n}\,, 
	\quad 
	\mathbf{x} \mapsto |\mathbf{x}\rangle = |x_1 x_2 \cdots x_n\rangle~.
\end{equation*}

This approach provides an intuitive one-to-one correspondence between classical bit strings and quantum states, enabling straightforward representation of discrete variables. However, it requires that all input features be represented in binary form prior to encoding. Moreover, basis encoding consumes one qubit per binary feature, limiting scalability when dealing with high-dimensional or real-valued datasets \citep{nielsen_quantum_2010}.

\textbf{Angle encoding}, also known as rotation encoding, represents classical data by assigning each component of a real-valued vector $\mathbf{x} \in \mathbb{R}^n$ to the rotation angle of a parameterized quantum gate, e.g. $\mathbf{R}_Y$, typically acting on a qubit initialized in the $|0\rangle$ state:

\vspace{-5pt}
\begin{equation*}
	\mathbb{R}^n \to \mathcal{H}_{2^n}\,, 
	\quad 
	\mathbf{x} \mapsto
	\Bigg(\bigotimes_{j=1}^{n} \mathbf{R}_Y(x_j)\Bigg)\,|0\rangle^{\otimes n}\,,
\end{equation*}
\vspace{-5pt}

where $\mathbf{R}_Y(x_j)$ denotes a rotation about the y-axis of the Bloch sphere by an angle proportional to the feature value $x_j$.
This maps continuous input data directly onto the Bloch sphere, with each qubit encoding one feature as a distinct quantum rotation \citep{havlicek_supervised_2019}.

Angle encoding offers a hardware-efficient way to represent continuous data and is widely used in \acp{VQC} due to its low resource requirements.
However, since each feature requires its own qubit and rotation gate, the representational capacity is linear in the number of available qubits, limiting its scalability compared to amplitude encoding.

\textbf{Amplitude encoding} is a data representation scheme that embeds the components of a classical vector into the amplitudes of a quantum state.
The idea dates back to early quantum algorithms such as the Harrow-Hassidim-Lloyd algorithm for solving linear systems \citep{harrow_quantum_2009}, and was formalized as a state-preparation problem in \citet{mottonen_transformation_2004} and \citet{plesch_quantum-state_2011}. 
In this scheme, the components of a classical feature vector are mapped directly to the probability amplitudes of the quantum system's basis states. 
Each amplitude corresponds to one possible realization of the quantum system, so that all information about the input vector is distributed across the superposition of states rather than stored in a single register element.
Since a quantum wavefunction represents a probability distribution over all possible basis states, its squared amplitudes must collectively sum to one, i.e.\ $\sum_i |x_i|^2 = 1$. 
To enforce this normalization condition for arbitrary real-valued vectors, the input is divided by its $\ell_2$-norm, ensuring that the resulting quantum state has unit total probability.

To illustrate, consider a two-qubit system: it spans a four-dimensional Hilbert space with computational basis states 
$|00\rangle$, $|01\rangle$, $|10\rangle$, and $|11\rangle$. 
Given an arbitrary real vector 
$\mathbf{x} \in \mathbb{R}^4$, 
its amplitude-encoded quantum representation is

\vspace{-8pt}
\begin{equation*}
|\boldsymbol{\psi}(\mathbf{x})\rangle 
= \frac{1}{\|\mathbf{x}\|_{\ell_2}}\big(x_1|00\rangle + x_2|01\rangle 
+ x_3|10\rangle + x_4|11\rangle\big)\,,
\quad 
\text{where} \quad
\|\mathbf{x}\|_{\ell_2}^2 = \sum_{i=1}^{4} x_i^2~.
\end{equation*}
\vspace{5pt}

This means that the quantum system simultaneously encodes all entries of the classical vector as part of a coherent superposition, where the probability of measuring each basis state equals the squared magnitude of its corresponding amplitude. 
The advantage of this encoding is that a single $n$-qubit register can represent $2^n$ classical features in parallel, in this case $4$ data points with only $2$ qubits. Amplitude encoding hence provides an exponentially compact data embedding and enables efficient quantum state manipulations.

Formally, any input vector $\mathbf{x} \in \mathbb{R}^{2^n}$ can be amplitude-encoded on an $n$-qubit quantum state as defined by

\vspace{-8pt}
\begin{gather}
	\mathbb{R}^{2^n}_{\setminus\{\mathbf{0}\}} \to \mathcal{H}_{2^n}, 
	\quad
	\mathbf{x} \mapsto
	\frac{1}{\|\mathbf{x}\|_{\ell_2}} \sum_{i=1}^{2^n} x_i\,|i\rangle,
	\quad
	\|\mathbf{x}\|_{\ell_2}^2 = \sum_{i=1}^{2^n} x_i^2~.
\label{eq:amplitude_encoding}
\end{gather}
\vspace{1pt}

To translate this mathematical idea into an actual quantum circuit, \citet{mottonen_transformation_2004} propose a state-preparation algorithm that constructs a \ac{PQC} capable of encoding any such amplitude-encoded quantum state. 
Their method decomposes the target state into a sequence of uniformly controlled rotations that iteratively align the amplitudes of computational basis states with the desired vector components. 
The circuit is constructed through a recursive sequence of uniformly controlled rotations---typically implemented with $\mathbf{R}_Y$ and $\mathbf{R}_Z$ gates for real and complex amplitudes, respectively, or more generally with arbitrary single-qubit unitaries $\mathbf{U}(\theta,\phi,\lambda)$---and conditional phase shifts, such that each rotation step adjusts the relative weights of the remaining amplitudes while preserving normalization. 
The number of single- and two-qubit gates in the resulting circuit scales with $\mathcal{O}(2^n)$ for an $n$-qubit register and achieves amplitude encoding of $\mathbf{x}$ onto the quantum state $|\boldsymbol{\psi}(\mathbf{x})\rangle$.

After encoding the classical data as a quantum state, in the second step of inference with a \ac{VQC} the parametrized unitary $\mathbf{U}(\boldsymbol{\theta})$ is constructed from alternating layers of single-qubit rotations and multi-qubit entangling gates.
Single-qubit rotations, such as $\mathbf{R}_X(\theta)$, $\mathbf{R}_Y(\theta)$, and $\mathbf{R}_Z(\theta)$, rotate individual qubits about the $x$, $y$, or $z$ axes of the Bloch sphere by angles determined by the corresponding parameters $\theta\in\boldsymbol{\theta}$.

\begin{figure}[H]
	\centering
	\includegraphics[width=0.65\textwidth]{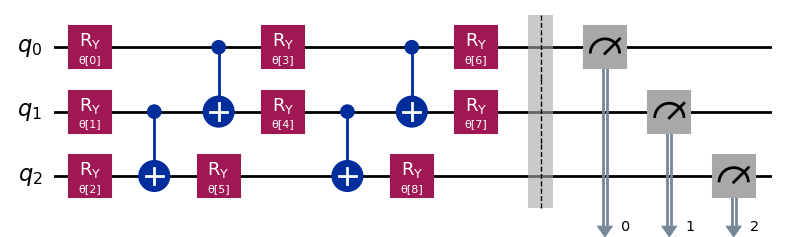}
	\caption{A PQC with three qubits, nine parameterized single-qubit rotation gates (red) and four two-qubit entangling CNOT gates (blue).}
	\label{fig:circuit}
\end{figure}

These unitary gates allow the circuit to explore arbitrary directions within the Hilbert space of each qubit.
Entangling layers, typically implemented with CNOT gates, couple the qubits and create non-classical correlations across them.
By stacking multiple layers of such rotation-entanglement blocks, \acp{PQC} can approximate increasingly expressive functions, analogous to the role of deeper layers in classical neural networks. Figure \ref{fig:circuit} shows an example of a simple \ac{PQC} with three qubits, three layers of rotations, two layers of entangling gates, and three measurement operations.

Once we construct the \ac{PQC} $\mathbf{U}(\boldsymbol{\theta})$ which includes input encoding and parameterized transformation, the final step is to extract the information from it. Generally, measurements in variational quantum algorithms are defined by Hermitian observables $\widehat{\bm{B}}$, whose expectation values encode task-relevant information about the quantum state. Given an observable $\widehat{\bm{B}}$ and input state $|\bm{x}\rangle$, a quantum circuit defines the expectation value:

\vspace{-8pt}
\begin{equation*}
    f(\bm{x}; \bm{\theta}) 
    = \langle \bm{x}| \, \mathbf{U}^\dagger(\bm{\theta}) \, \widehat{\bm{B}} \, \mathbf{U}(\bm{\theta}) \, |\bm{x} \rangle.
    \label{eq:vqc}
\end{equation*}

Common choices for observables include Pauli operators and their tensor products, among which, the Pauli-$Z$ operator is particularly convenient, as it corresponds to measurement in the computational basis and directly quantifies the position on the z-axis of the Bloch sphere, that is, the population imbalance between $\lvert 0 \rangle$ and $\lvert 1 \rangle$. 
This makes Pauli-$Z$ expectation values naturally interpretable as the probability of measuring $\lvert 1 \rangle$, and they are therefore widely used as readout observables in variational quantum circuits.

However, quantum measurements are inherently probabilistic: measuring a qubit yields a binary outcome, corresponding to a collapse onto either $\lvert 0 \rangle$ or $\lvert 1 \rangle$. 
As a result, expectation values cannot be obtained from a single circuit execution, but must instead be estimated empirically by repeating the circuit multiple times (shots). 
The repeated sampling of the \ac{PQC} yields an estimate of the probability distribution over measurement outcomes.
The expectation values of specific observables (averages of the observed outcomes) constitute the model outputs.

In the context of variational algorithms, these measured quantities are then used to evaluate a real-valued cost function $C(\boldsymbol{\theta})$ that quantifies the deviation between the model predictions and the desired target values.
The parameters $\boldsymbol{\theta}$ are updated based on this cost, forming a closed-loop optimization process akin to classical neural networks.

This optimization requires gradients of expectation values with respect to the parameters $\bm{\theta}$. 
\citet{mitarai_quantum_2018} demonstrate that, for parametrized quantum gates of the form $\mathbf{U}(\theta) = e^{-i \, \theta \mathbf{G}/2}$ where $\mathbf{G}$ is Hermitian with two distinct eigenvalues, the gradient of an observable expectation value can be evaluated analytically via what is now known as the \emph{parameter-shift rule}:

The derivative of $f$ with respect to a single parameter is given by

\vspace{-5pt}
\begin{equation}
    \frac{\partial}{\partial \theta_i} f(\bm{x}; \bm{\theta}) 
    = \tfrac{1}{2} \left[ 
        f\!\left(\bm{x}; \bm{\theta} + \tfrac{\pi}{2}\hat{e}_i \right)
        - f\!\left(\bm{x}; \bm{\theta} - \tfrac{\pi}{2}\hat{e}_i \right) 
      \right],
\label{eq:param_shift}
\end{equation}
\vspace{-5pt}

where $\hat{e}_i$ denotes the unit vector in the $i$th coordinate direction.

Since this condition holds for single-qubit rotation gates such as $\mathbf{R}_X(\theta)$, $\mathbf{R}_Y(\theta)$, and $\mathbf{R}_Z(\theta)$, the rule can be directly applied to the trainable layers of a \ac{VQC} to compute exact gradients of the circuit output with respect to its parameters.
In other words, the gradient of the expectation value can be calculated by evaluating the same VQC twice per parameter, with the $i$-th parameter shifted by $\pm \tfrac{\pi}{2}$ while keeping all others fixed. This enables unbiased gradient estimation directly from quantum measurements \citep{schuld_evaluating_2019}.

\begin{figure}[H]
    \centering
    \includegraphics[width=\textwidth]{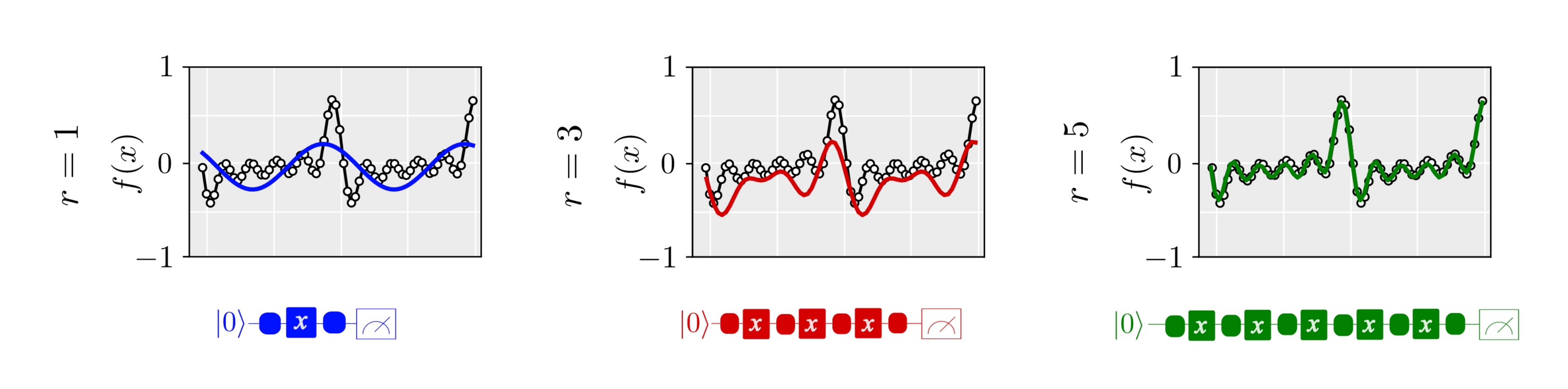}
	\caption{Visualization of converged \acp{VQC} with one, three, and five layers of rotation and entanglement gates, illustrating how increasing circuit depth enables the representation as higher-order Fourier series. \citep[Adapted from][]{schuld_effect_2021}.}
	\label{fig:fourier}	
\end{figure}

By iteratively adjusting the parameter vector $\boldsymbol{\theta}$ to minimize the cost function $C(\boldsymbol{\theta})$, the VQC converges to a high-order Fourier series representation of the provided input data, as illustrated in Figure \ref{fig:fourier}. It therefore acts as an \emph{asymptotically universal} approximator.

\vspace{-10pt}

The expressive power of VQCs lies in their ability to generate highly entangled states over exponentially large Hilbert spaces, but practical training suffers from the phenomenon of barren plateaus, in which gradients vanish exponentially with system size and circuit depth \citep{mcclean_barren_2018}. This imposes strong constraints on circuit design, initialization, and ansatz selection, motivating problem-inspired architectures and shallow circuits adapted to the NISQ regime, like the one we propose in this work.

\subsection{Reinforcement Learning}

Reinforcement Learning (RL) is a computational framework for learning behaviour through trial-and-error interaction with a dynamic environment.
At each discrete time step $t$, the agent observes a state $\mathbf{s}_t$, selects an action $\mathbf{a}_t$, receives a reward $r_t$, and transitions to a new state $\mathbf{s}_{t+1} \sim P(\cdot \mid \mathbf{s}_t, \mathbf{a}_t)$.
The probability of taking action $\mathbf{a}$ in state $\mathbf{s}$ is defined by the agent's policy $\pi(\mathbf{a} \mid \mathbf{s})$.
The optimal policy $\pi^*$ is defined as the one that maximizes the Bellman equations $V^\pi(\mathbf{s})$ and $Q^\pi(\mathbf{s},\mathbf{a})$ for the entire state-action space \citep{bellman_dynamic_1957}.
\begin{align*}
V^\pi(\mathbf{s}) &= \mathbb{E}_{\mathbf{a} \sim \pi(\cdot|\mathbf{s})} \!\left[
R(\mathbf{s},\mathbf{a}) + \gamma \mathbb{E}_{\mathbf{s}' \sim P(\cdot|\mathbf{s},\mathbf{a})} V^\pi(\mathbf{s}')
\right], \\[6pt]
Q^\pi(\mathbf{s},\mathbf{a}) &= R(\mathbf{s},\mathbf{a}) + \gamma \mathbb{E}_{\mathbf{s}' \sim P(\cdot|\mathbf{s},\mathbf{a})} 
\mathbb{E}_{\mathbf{a}' \sim \pi(\cdot|\mathbf{s}')} Q^\pi(\mathbf{s}',\mathbf{a}').
\end{align*}

While the Bellman equations characterize optimal behaviour, they do not by themselves provide a practical solution in most settings.
Solving them exactly requires either full knowledge of the environment dynamics or exhaustive enumeration of the state-action space, both of which quickly become intractable as problem complexity grows.
\ac{RL} algorithms can be viewed as different strategies for approximately solving one of the Bellman optimality equations under varying assumptions.

We summarize the algorithms most relevant for contextualizing our methodological framework, but there exists a wide range of approaches to solve this problem.
For a comprehensive treatment of \ac{RL} and its origins, we refer the reader to \cite{sutton_reinforcement_2015}. 

\subsubsection*{Q-Learning and Deep Q-Networks}

Q-learning \citep{watkins_q-learning_1992} is an off-policy \ac{TD} control algorithm that learns the optimal action-value function independently of the behaviour policy. Unlike on-policy methods, Q-learning updates action values toward the maximum action-value of the next state, regardless of which action is actually taken, as formalized in Algorithm~\ref{alg:ql}. This property allows Q-learning to learn optimal policies even when following exploratory or suboptimal behaviour policies, making it a powerful and widely used RL algorithm.

\begin{algorithm}[H]
  \caption{Q-Learning}
  \begin{algorithmic}[1]
    \State Initialize action-value function $Q(\mathbf{s},\mathbf{a})$ arbitrarily
    \For{each episode}
      \State Initialize $\mathbf{s}$
      \Repeat
        \State Select $\mathbf{a}$ via $\varepsilon$-greedy w.r.t. $Q(\mathbf{s},\cdot)$
        \State Execute $\mathbf{a}$, observe $r$, $\mathbf{s}'$
        \State $Q(\mathbf{s},\mathbf{a}) \leftarrow Q(\mathbf{s},\mathbf{a}) + \alpha \left[ r + \gamma \max_{\mathbf{a}'} Q(\mathbf{s}',\mathbf{a}') - Q(\mathbf{s},\mathbf{a}) \right]$
        \State $\mathbf{s} \leftarrow \mathbf{s}'$
      \Until{terminal condition}
    \EndFor
  \end{algorithmic}
  \label{alg:ql}
\end{algorithm}

Action selection is implemented using the $\varepsilon$-greedy policy: with probability $1-\varepsilon$, the agent selects the action with the highest estimated action value, while with probability $\varepsilon$ it selects an action uniformly at random.
This simple exploration strategy ensures sufficient state-action space coverage while maintaining a bias toward actions that are currently estimated to be optimal.
While tabular Q-learning converges under suitable conditions, it also becomes computationally infeasible in large state-action spaces where the action-value function cannot be stored explicitly.

\acp{DQN} \citep{mnih_human-level_2015} address this limitation by approximating $Q(s,a)$ with a neural network. To stabilize training, DQNs employ experience replay and a separate target network, which may be updated periodically or via soft updates.
The resulting training procedure is delineated in Algorithm~\ref{alg:dqn}. These extensions enable Q-learning to scale to high-dimensional observation spaces and form the foundation of modern value-based Deep \ac{RL}.

However, the reliance on an explicit maximization over actions (see Line 12 of Algorithm \ref{alg:dqn}) still restricts \acp{DQN} to discrete action spaces.
In continuous settings, this maximization can only be approximated by sampling a finite set of candidate actions and selecting the best among them.
This limitation motivates Actor-Critic methods, which learn explicit policies and enable direct action sampling together with more flexible exploration mechanisms.

\begin{algorithm}[H]
  \caption{Deep Q-Network (DQN)}
  \begin{algorithmic}[1]
    \State Initialize action-value network \( Q(\mathbf{s},\mathbf{a} \mid \bm{\theta}^Q) \) arbitrarily
    \State Initialize target network \( Q'(\mathbf{s},\mathbf{a} \mid \bm{\theta}^{Q'}) \gets Q(\mathbf{s},\mathbf{a} \mid \bm{\theta}^Q) \)
    \State Initialize replay buffer \( \mathcal{D} \)
    \State Initialize target network update period $C$
    \For{each episode}
      \State Receive initial state \( \mathbf{s} \)
      \For{each step}
        \State Select action \( \mathbf{a} \) via $\varepsilon$-greedy w.r.t. $Q(\mathbf{s},\cdot)$
        \State Execute \( \mathbf{a} \), observe reward \( r \) and next state \( \mathbf{s}' \)
        \State Store transition \( (\mathbf{s},\mathbf{a},r,\mathbf{s}') \) in \( \mathcal{D} \)
        \State Sample minibatch \( \{(\mathbf{s}_j,\mathbf{a}_j,r_j,\mathbf{s}'_j)\}_{j=1}^N \sim \mathcal{D} \)
        \State Compute target: $y_j \gets r_j + \gamma \max_{\mathbf{a}'} Q'(\mathbf{s}'_j, \mathbf{a}' \mid \bm{\theta}^{Q'})$
        \State Update $\bm{\theta^Q}$ to minimize MSE loss: $\mathcal{L}(\bm{\theta^Q}) = \frac{1}{N} \sum_{j}^N \big(Q(\mathbf{s}_j, \mathbf{a}_j \mid \bm{\theta}^Q) - y_j\big)^2 $
        \If{step mod $C = 0$}
            \State Update target network: \( \bm{\theta}^{Q'} \gets \bm{\theta}^Q \)
        \EndIf
        \State Set \( \mathbf{s} \gets \mathbf{s}' \)
        \EndFor
    \EndFor
  \end{algorithmic}
  \label{alg:dqn}
\end{algorithm}

\subsubsection*{Deep Deterministic Policy Gradient (DDPG)}

\ac{DDPG} \citep{lillicrap_continuous_2019} extends deterministic policy gradient methods to deep neural networks and continuous action spaces.
In contrast to stochastic Actor-Critic algorithms, \ac{DDPG} employs a deterministic actor $\mu(\mathbf{s} \mid \bm{\theta}^\mu)$ that maps each state directly to a continuous action, together with a critic $Q(\mathbf{s}, \mathbf{a} \mid \bm{\theta}^Q)$ that estimates the corresponding action-value function.
Training is performed off-policy, leveraging experience replay, target networks, and soft updates to stabilize learning.

The target value $y$ is computed through the target networks $\mu'(\mathbf{s}' \mid \bm{\theta}^{\mu'})$ and $Q'(\mathbf{s}', \mathbf{a}' \mid \bm{\theta}^{Q'})$:
\[
y = r + \gamma Q'(\mathbf{s}', \mu'(\mathbf{s}' \mid \bm{\theta}^{\mu'}) \mid \bm{\theta}^{Q'})\,,
\]

and the critic parameters are updated by minimizing the mean-squared error (MSE) loss
\[
\mathcal{L}(\bm{\theta}^Q) = \mathbb{E}_{(\mathbf{s},\mathbf{a},r,\mathbf{s}') \sim \mathcal{D}}
\big[ \big( Q(\mathbf{s},\mathbf{a} \mid \bm{\theta}^Q) - y \big)^2 \big],
\]

where $\mathcal{D}$ denotes the replay buffer.

The actor is updated using the deterministic policy gradient, which propagates gradients through the critic with respect to the action. This update adjusts the actor parameters to maximize the critic's estimated action value:
\[
\nabla_{\theta^\mu} J \approx 
\mathbb{E}_{(\mathbf{s},\mathbf{a})\sim\mathcal{D}} 
\left[ 
\nabla_a Q(\mathbf{s},\mathbf{a} \mid \bm{\theta}^Q)\big|_{\mathbf{a}=\mu(\mathbf{s})} 
\nabla_{\bm{\theta}^\mu} \mu(\mathbf{s})
\right].
\]

Both the actor and the critic are trained using minibatches sampled from $\mathcal{D}$, and the target networks are updated via soft updates.
The complete \ac{DDPG} procedure is outlined in Algorithm~\ref{alg:ddpg}.
Together, these design choices enable stable and efficient learning in high-dimensional continuous control problems.

\begin{algorithm}[H]
  \caption{DDPG}
  \begin{algorithmic}[1]
    \State Initialize actor \( \mu(\mathbf{s}\mid\bm{\theta}^\mu) \) and critic \( Q(\mathbf{s},\mathbf{a}\mid\bm{\theta}^Q) \) network parameters arbitrarily
    \State Initialize target network parameters $\bm{\theta}^{\mu'} \gets \bm{\theta}^\mu$, $\bm{\theta}^{Q'} \gets \bm{\theta}^Q$
    \State Initialize replay buffer $\mathcal{D}$
    \For{each episode}
      \State Observe initial state $\mathbf{s}$
      \Repeat
        \State Select action $\mathbf{a} \gets \mu(\mathbf{s}) + \varepsilon$, $\varepsilon \sim \mathcal{N}(0, \sigma^2)$
        \State Execute $\mathbf{a}$, observe reward $r$ and next state $\mathbf{s}'$
        \State Store transition \( (\mathbf{s},\mathbf{a},r,\mathbf{s}') \) in \( \mathcal{D} \)
        \State Sample minibatch \( \{(\mathbf{s}_j,\mathbf{a}_j,r_j,\mathbf{s}'_j)\}_{j=1}^N \sim \mathcal{D} \)
        \State Compute target $y = r + \gamma Q'(\mathbf{s}', \mu'(\mathbf{s}' \mid \bm{\theta}^{\mu'}) \mid \bm{\theta}^{Q'})$
        \State Update critic by minimizing $\big(Q(\mathbf{s},\mathbf{a} \mid \bm{\theta}^Q) - y\big)^2$
        \State Update actor:
        \( 
        \nabla_{\bm{\theta}^\mu} J \approx 
        \frac{1}{N} \sum_j^N
        \nabla_a Q(\mathbf{s}_j, \mathbf{a} \mid \bm{\theta}^Q)\big|_{\mathbf{a}=\mu(\mathbf{s}_j)} 
        \nabla_{\bm{\theta}^\mu} \mu(\mathbf{s}_j)
        \)
        \State Soft update target networks:
        $\bm{\theta}^{Q'} \gets \tau \bm{\theta}^Q + (1{-}\tau)\bm{\theta}^{Q'}$, 
        $\bm{\theta}^{\mu'} \gets \tau \bm{\theta}^\mu + (1{-}\tau)\bm{\theta}^{\mu'}$
        \State $\mathbf{s} \gets \mathbf{s}'$
      \Until{terminal condition}
    \EndFor
  \end{algorithmic}
  \label{alg:ddpg}
\end{algorithm}

\subsection{Reinforcement Learning for Portfolio Optimization}

Portfolio optimization is inherently a sequential decision-making problem.
Investment decisions are not made in isolation at a single point in time, but rather as part of a dynamic process in which the portfolio is rebalanced within certain time intervals.
Market conditions evolve over time, rewards are often delayed, and actions taken today constrain or enable actions tomorrow.
These characteristics are difficult to capture with static optimization techniques that treat portfolio construction as a one-shot problem.

\ac{RL} provides a suitable framework for addressing these issues by casting portfolio management as a sequential control problem, incorporating delayed rewards, state transitions, and the exploration-exploitation tradeoff.
Formally, we express portfolio optimization as a dynamic decision-making problem over a state-action space, in which the agent learns a trading policy that maps market states to portfolio allocations in order to maximize a risk-adjusted performance objective over time.

\vspace{-9pt}
\begin{align}
\label{eq:state_action_space}
\text{State:} \quad &\mathbf{s}_t = \big[\, \mathbf{p}_{t-L:t},\; \hat{\mathbf{p}}_{t+1:t+F+1} \,\big], 
\quad 
\text{Action:} \quad \mathbf{a}_t = \mathbf{w}_t\,, \\[8pt]\nonumber
\text{Reward:} \quad &
r_t = 
\mathbf{w}_t^\top \boldsymbol{\mu}_{t-L:t+F+1}
- \eta\, \mathbf{w}_t^\top \boldsymbol{\Sigma}_{t-L:t+F+1}\, \mathbf{w}_t\,, \\[8pt]\nonumber
\text{Objective:} \quad &\hspace{-2.2pt}\max_{\mathbf{a}_t} ~\, Q_t(\mathbf{s}_t, \mathbf{a}_t) =
\Big[
r_t(\mathbf{s}_t, \mathbf{a}_t)
+ \gamma \,
\mathbb{E} \big[ Q_{t+1}(\mathbf{s}_{t+1}, \mathbf{a}_{t+1}) \big]
\Big], \\[8pt]\nonumber
\text{where} \quad
& \mathbf{w}_t \in \mathbb{R}^N ~\text{: portfolio weight vector (allowing for short-selling)}, \\[4pt]\nonumber
& \mathbf{p}_t \in \mathbb{R}^N ~\text{: vector of past asset prices}, \\[4pt]\nonumber
& \hat{\mathbf{p}}_t \in \mathbb{R}^N ~\text{: vector of forecasted asset prices}, \\[4pt]\nonumber
& \eta,\,\gamma \in [0,1] ~\text{: risk preference parameter and Bellman equation discount factor}, \\[4pt]\nonumber
& L,\, F \in \mathbb{N} ~\text{: lookback and forecast window size in days}, \nonumber \\[4pt]
& N,\, t \in \mathbb{N} ~\text{: number of assets and timestep (day)}. \nonumber
\end{align}

The action corresponds to the portfolio allocation vector at each time step, represented as continuous weights across all assets. The state is constructed from a rolling window of the past daily prices for each asset, concatenated with a forecast, thereby capturing both historical dynamics and short-term predictive information. The reward balances profitability and risk through a combination of average returns and risk-adjusted volatility with a tunable risk preference parameter $\eta$, mirroring the quadratic optimization problem.

This formulation integrates risk directly into the reward signal, avoiding the division by risk inherent in Sharpe ratio-based objectives, which can induce uneven loss landscapes and numerical instabilities during training.

Value-based methods such as Q-learning and \acp{DQN} rely exclusively on approximating value functions and derive policies indirectly through action-value maximization. While conceptually simple, these approaches become impractical in continuous action spaces without additional approximations, such as action discretization or sampling-based maximization. Moreover, even accurate value-function approximation does not guarantee that the induced policy is near-optimal when function approximation errors are present.

Policy gradient methods take the opposite approach by directly optimizing a parameterized policy. This allows them to operate naturally in continuous action spaces and optimize the control objective explicitly. However, because policy gradients are estimated from sampled trajectories, they often suffer from high variance and strong sensitivity to hyperparameters. In addition, successive gradient estimates are typically computed independently, limiting the accumulation and reuse of information across iterations.

Actor-Critic architectures combine these two perspectives by coupling a policy (the actor) with a value function estimator (the critic). The critic provides a learned baseline that reduces the variance of policy gradient updates, while the actor ensures direct optimization in policy space. This interaction enables more stable learning and improved scalability to high-dimensional state-action spaces, making Actor-Critic methods particularly well suited to financial decision-making problems like portfolio optimization \citep{konda_actor-critic_1999}.

We therefore adopt \ac{DDPG} and a continuous-action variant of a \ac{DQN} with a parameterized actor as the basis for our RL approach.
Both methods are designed for continuous action domains such as portfolio optimization, where actions correspond to fractional asset allocations.
The key difference between the two algorithms is, that \ac{DDPG} constructs its targets using the next action proposed by the learned actor, while the continuous \ac{DQN} instead approximates the action-value maximization by sampling a finite set of possible next actions from the learned actor and selecting the action-value maximizing action.
Both approaches are trained off-policy, allowing extensive reuse of historical experience through replay buffers and stabilizing updates via target networks and soft parameter updates, facilitating data efficiency and robust learning.

The complete training procedures are provided in Algorithm~\ref{alg:classical}, which details initialization, replay buffer usage, policy updates, and the respective target computation strategies.

\begin{algorithm}[H]
    \caption{DDPG/DQN for Portfolio Optimization}
    \begin{algorithmic}[1]
        \State Initialize actor \( \mu(\mathbf{s}\mid\bm{\theta}^\mu) \) and critic \( Q(\mathbf{s},\mathbf{a}\mid\bm{\theta}^Q) \) parameters
        \State Initialize target network parameters: \( \bm{\theta}^{\mu'} \gets \bm{\theta}^\mu \), \( \bm{\theta}^{Q'} \gets \bm{\theta}^Q \)
        \State Initialize replay buffer \( \mathcal{D} \)
        \For{episode $= 1$ to $M$}
            \State Receive initial state \( \mathbf{s} \)
            \For{$t = 1$ to $T$}
                \State Select action: \( \mathbf{a} \gets \mu(\mathbf{s}\mid\bm{\theta}^\mu) + \varepsilon \), where \( \varepsilon \sim \mathcal{N}(0, \sigma^2) \)
                \State Execute \( \mathbf{a} \), observe reward \( r \) and next state \( \mathbf{s}' \)
                \State Store transition \( (\mathbf{s},\mathbf{a},r,\mathbf{s}') \) in \( \mathcal{D} \)
                \State Sample minibatch \( \{(\mathbf{s}_j,\mathbf{a}_j,r_j,\mathbf{s}'_j)\}_{j=1}^N \sim \mathcal{D} \)
                \If{DDPG}
                    \State Compute target: \( y_j \gets r_j + \gamma Q'(\mathbf{s}'_j, \mu'(\mathbf{s}'_j) \mid \bm{\theta}^{Q'}) \)
                \ElsIf{DQN}
                    \State Sample \( \{\mathbf{a}'_{j,k}\}_{k=1}^M \sim \text{Uniform}(\mathcal{A}) \)
                    \State Compute target: \( y_j \gets r_j + \gamma \max_{k} Q'(\mathbf{s}'_j, \mathbf{a}'_{j,k} \mid \bm{\theta}^{Q'}) \)
                \EndIf
                \State Update critic: minimize \( \mathcal{L}(\bm{\theta}^Q) = \frac{1}{N} \sum_j^N (Q(\mathbf{s}_j, \mathbf{a}_j \mid \bm{\theta}^Q) - y_j)^2 \)
                \State Update actor: \( \nabla_{\bm{\theta}^\mu} J \approx \frac{1}{N} \sum_j^N \nabla_a Q(\mathbf{s}_j, \mathbf{a} \mid \bm{\theta}^Q) \big|_{\mathbf{a}=\mu(\mathbf{s}_j)} \cdot \nabla_{\bm{\theta}^\mu} \mu(\mathbf{s}_j) \)
                \State Soft update target networks: \( \bm{\theta}^{Q'} \gets \tau \bm{\theta}^Q + (1{-}\tau) \bm{\theta}^{Q'} \), \quad \( \bm{\theta}^{\mu'} \gets \tau \bm{\theta}^\mu + (1{-}\tau) \bm{\theta}^{\mu'} \)
                \State Set \( \mathbf{s} \gets \mathbf{s}' \)
            \EndFor
        \EndFor
    \end{algorithmic}
	\label{alg:classical}
\end{algorithm}

In the following section, we detail how we translate the above classical algorithm into a fully quantum framework based on \acp{VQC}, allowing for a controlled comparison between classical and quantum RL in the context of the portfolio optimization problem.

\subsection{Quantum Reinforcement Learning for Portfolio Optimization}

We adapt two canonical reinforcement learning algorithms to a fully quantum setting: \ac{DDPG} and Deep Q-Learning.
By unifying both classical and quantum approaches within the same \ac{RL} frameworks, we enable a systematic comparison of their effectiveness for portfolio optimization while demonstrating the flexibility of \acp{VQC} in supporting distinct reinforcement learning methodologies.
Both methods are implemented in continuous action space, where the action corresponds to a portfolio allocation vector, and both employ an experience replay buffer and target networks to stabilize learning. The architectures are structurally similar, with actor and critic networks realized as \acp{PQC}, and training driven by the same interaction loop between agent and environment as described in Algorithm \ref{alg:classical} in the previous section.
The outputs of the circuits defined as the expectation values of observables $\widehat{\bm{B}}$ acting on the $N$ first qubits.

\vspace{-5pt}
\begin{equation}
	f_{\mathrm{VQC}}(\mathbf{x}; \bm{\theta}) 
	= \langle \mathbf{x}| \mathbf{U}^\dagger(\bm{\theta}) \, \widehat{\bm{B}} \, \mathbf{U}(\bm{\theta}) |\mathbf{x} \rangle
\end{equation}
\vspace{-5pt}

We map classical financial observations onto the quantum domain by constructing a normalized feature map and encoding it as a quantum state via amplitude encoding.
To this end, we extend the canonical amplitude encoding defined in Equation \ref{eq:amplitude_encoding} by constructing a normalized feature map that transforms the input vector before embedding. 
Instead of directly encoding the normalized input $\tilde{\mathbf{x}}$, we transform it into a higher-dimensional feature vector $\mathbf{Z}(\tilde{\mathbf{x}})$ that captures both linear and nonlinear dependencies among components of the input. 
The resulting amplitude-encoded quantum state is defined as

\vspace{-10pt}
\begin{gather*}
|\boldsymbol{\psi}(\mathbf{Z})\rangle 
= \frac{1}{\|\mathbf{Z}\|_{\ell_2}} \sum_{i=1}^{d} z_i\,|i\rangle,\quad d = \text{dim}(\mathcal{H}), \\[6pt]
\text{where} \quad
\mathbf{Z}(\tilde{\mathbf{x}}) 
= \left[\, \tilde{\mathbf{x}},\ \tilde{\mathbf{x}}^{\odot 2},\ 
          \sin(\tilde{\mathbf{x}}),\ \cos(\tilde{\mathbf{x}})\, \right],\\[8pt]
\text{and} \quad
\tilde{\mathbf{x}}
= \frac{\mathbf{x} - \frac{1}{n}\sum_{i=1}^n x_i}
       {\sqrt{\tfrac{1}{n}\sum_{i=1}^n \left(x_i - \tfrac{1}{n}\sum_{j=1}^n x_j\right)^2}}~.
\end{gather*}
\vspace{1pt}

By applying amplitude encoding to the extended feature map $\mathbf{Z}$ rather than to the normalized input $\tilde{\mathbf{x}}$, we break the radial symmetry in the input space caused by the division by the norm.
This enhances expressivity and helps to mitigate barren plateaus during training, as the circuit can more easily differentiate between distinct input states.

To extract a portfolio allocation from the actor circuit, we adopt a measurement approach aligned with the asset dimension. Given a state--action space with $N$ assets, we measure the first $N$ qubits of the circuit in the computational basis using repeated shots, yielding empirical marginal probabilities $\mathbf{p} \in [0,1]^N$ for observing the state $\lvert 1 \rangle$ on each qubit. These probabilities are obtained as expectation values of single-qubit Pauli-$Z$ observables. To obtain a valid portfolio allocation, the resulting vector is $\ell_1$-normalized to produce portfolio weights $\mathbf{w} \in \mathbb{R}^N$ satisfying $\sum_i w_i = 1$. This normalization enforces full capital allocation while allowing both long and short positions, with positive and negative weights corresponding to long and short exposures, respectively.

\begin{align}
{\mathbf{w}}(\mathbf{x};\bm{\theta}) 
&= \mathcal{N}\!\left(
\langle \mathbf{x}| \mathbf{U}^\dagger(\bm{\theta}) \,
\widehat{\mathbf{B}} \,
\mathbf{U}(\bm{\theta}) |\mathbf{x} \rangle
\right), \\[6pt]\nonumber
\text{where} \qquad
\widehat{\mathbf{B}} = (Z_1,\dots,&Z_N)\,,
\qquad
\mathcal{N}(\mathbf{z})_i
= \frac{z_i}{\sum_{j=1}^{N} z_j},
\quad i = 1,\dots,N\,.
\end{align}

To compute gradients of expectation values with respect to circuit parameters, we employ the parameter-shift rule described in Equation~\ref{eq:param_shift}.
This technique enables exact and unbiased gradient estimation for variational circuits composed of parameterized rotation gates, without requiring analytical differentiation or finite-difference approximations.
Each parameter update involves two forward evaluations of the circuit with fixed parameter shifts, yielding an exact derivative for every trainable parameter $\theta_i$.

The resulting gradients $\nabla_{\bm{\theta}} f_{\mathrm{VQC}}(\mathbf{x}; \bm{\theta})$ are fully compatible with classical optimization routines.
They can be directly integrated into standard backpropagation pipelines, where the quantum circuit is treated as a differentiable module within the computational graph.
This allows the use of conventional gradient-based optimizers such as stochastic gradient descent or Adam \citep{kingma_adam_2017} to iteratively minimize a real-valued loss function defined on the quantum circuit outputs.

\begin{figure}[H]
    \centering
    \includegraphics[width=1.02\textwidth]{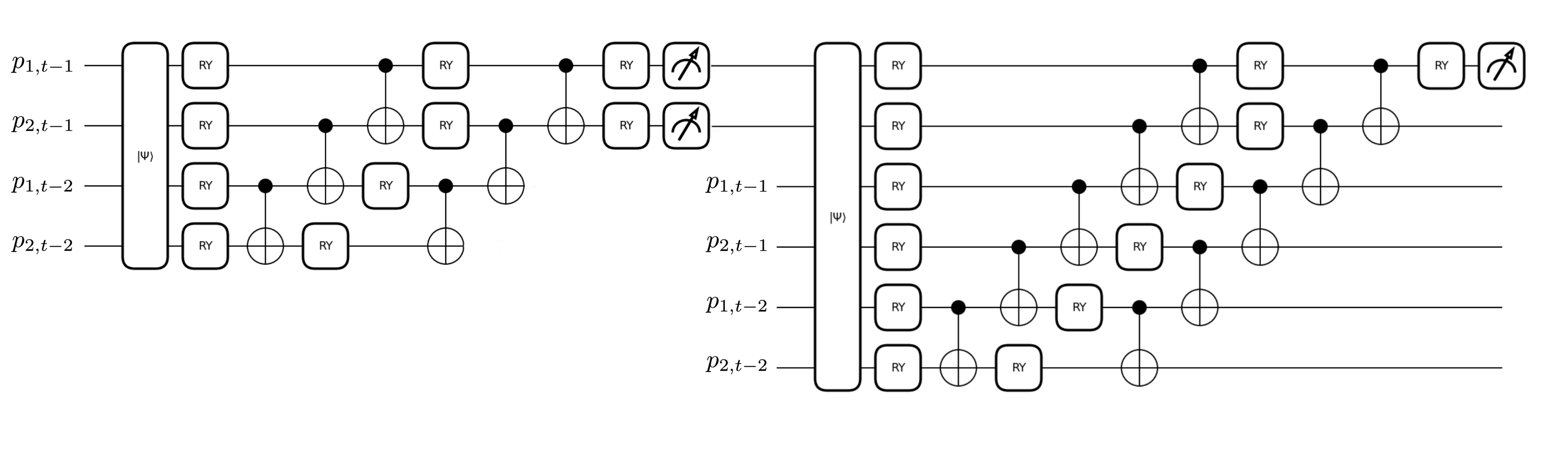}
    \caption{Architecture of our proposed \ac{QRL} pipeline for portfolio optimization in the illustrative case of a two-asset, two-time-step environment.}
    \label{fig:architecture}
\end{figure}

Unlike hybrid models that retain a classical function approximator (e.g. a neural network) side by side with the VQC, our formulation is \emph{fully quantum}: both actor and critic are represented directly in Hilbert space. Only the calculation of the loss and backpropagation are conducted on classical hardware. This architecture provides a quantum-native instantiation of reinforcement learning, allowing us to assess the representational benefits of variational circuits while keeping the training dynamics comparable to classical baselines. Figure~\ref{fig:architecture} depicts this architecture schematically for the illustrative case of two assets and two historical timesteps.

Each horizontal wire corresponds to a qubit. Classical input features (lagged market price information such as $p_{i,t-1}$ and $p_{i,t-2}$) are embedded into the quantum state $\lvert \psi \rangle$ via normalized amplitude embedding, forming the data-encoding layer of the circuit. This encoding maps the classical state of the environment into a coherent quantum representation while remaining hardware efficient and scalable.

Following encoding, the circuit consists of alternating layers of parameterized single-qubit rotations and structured entangling gates. The entanglement pattern is designed to couple both assets and temporal dimensions, enabling the circuit to learn cross-asset dependencies as well as intertemporal relationships. The repeated variational layers constitute the trainable core of both the actor and critic portions of the circuit. In the actor circuit, the final measurement layer extracts Pauli-$Z$ expectation values from a subset of qubits, which are classically normalized to produce a continuous portfolio allocation satisfying the full-investment constraint. In the critic circuit, an analogous readout produces a scalar output in the range $[-1,1]$, which is interpreted as an estimate of the risk-adjusted portfolio performance, corresponding to the Sharpe ratio associated with the actor's proposed allocation.

The example shown in Figure~\ref{fig:architecture} scales directly to larger asset universes and longer observation windows without changing the underlying architecture. Scalability is governed by the amplitude-encoding stage described earlier. Increasing the number of assets or timesteps increases the classical input dimension, which is first expanded by the feature map and then embedded into a quantum state $\lvert \psi \rangle$ using a logarithmic number of qubits. Since the feature map expands the input only by a constant factor and amplitude encoding requires $\mathcal{O}(\log d)$ qubits for a $d$-dimensional input, the circuit width grows logarithmically as the effective feature space scales linearly, making the experimental design described in the following section feasible.


\section{Experimental Design}

In the following we describe the experimental design used to assess the effectiveness of the proposed \ac{QRL} framework in a realistic portfolio optimization setting.
Rather than relying on idealized or synthetic benchmarks, we ground our evaluation in historical financial data and formulate portfolio allocation as a continuous control problem that reflects practical investment constraints.
States, actions, and rewards are defined to align with standard portfolio management practice, enabling a meaningful comparison between learning-based strategies.

To contextualize the performance of the quantum agents, we benchmark them against static classical baselines and state-of-the-art Deep \ac{RL} methods operating under identical market conditions.
This experimental setup is intended to isolate the impact of the quantum policy representation while ensuring that observed performance differences arise from the learning architecture itself.

\subsection{Data and Environment}

The dataset consists of 5049 daily closing prices for 15 diverse financial assets, including equities, fixed income instruments, and real assets. Asset returns are computed as logarithmic differences.
The full list of assets is provided in Appendix \ref{app:assets}.
The data span a period from August 2011 to September 2025, covering multiple market cycles and regimes.

Each environment state comprises a 30-day rolling window of normalized asset prices concatenated with a 7-day forecast obtained using an Auto ARIMA model \citep{box_time_1970}, allowing the agent to condition its decisions on predicted short-term market movements in addition to past returns.
This hybrid state representation captures both short-term momentum and expected trend behaviour, allowing the agent to condition its policy on recent dynamics as well as forward-looking signals.

All \ac{RL} approaches considered in this work are inherently dynamic and forward-looking.
Portfolio allocations are dynamically rebalanced at a fixed frequency of 30 days.
Forward-looking information is incorporated implicitly through the optimization of expected future rewards and explicitly through the forecast window added to the state representation.

Actions are continuous-valued portfolio allocations, where negative values correspond to short positions.
At each decision point, the agent outputs a 15-dimensional vector $\mathbf{w}_t$, where $w_{t,i}$ denotes the proportion of capital allocated to asset $i$ at time $t$. The portfolio is rebalanced every 30 days.
Short selling is permitted up to 100 \% of the initial portfolio value, reflecting realistic leverage constraints in institutional asset management. Transitions in the environment are governed by observed market returns.

\subsection{Objective and Evaluation Metrics}

The main performance metric used for evaluation in this study is the Sharpe ratio of the resulting portfolio on the held-out test segment, averaged across the seven folds of our increasing rolling-window time-series cross-validation.
All results are computed while taking into account a transaction cost of 0.15 \% per trade, corresponding to the transaction costs of a typical large institutional investor \citep{frazzini_trading_2018}.

The Sharpe ratio is computed as the ratio of excess return to volatility, adjusted with a small constant $\varepsilon$ in the denominator to prevent division by zero.

\vspace{-5pt}
\begin{align*}
	\text{Sharpe Ratio} &= \frac{R_p - R_f}{\sigma_p + \varepsilon}, \\[6pt]
	R_p &= \frac{1}{T} \sum_{t=1}^{T} R_{p,t}, \quad R_{p,t} = \sum_{i=1}^{N} p_{t,i} \left(w_{t,i} - \frac{0.15}{100} \left| w_{t,i} - w_{t-1,i} \right|\right), \\[7pt]
	R_f &= 0.0418 \quad \text{(US 10-Year Treasury yield)}, \\[6pt]
	\sigma_p &= \sqrt{\frac{1}{T - 1} \sum_{t=1}^{T} (R_{p,t} - R_p)^2}, \quad \varepsilon = 10^{-7}.
\end{align*}

\begin{figure}[H]
	\centering
	{\includegraphics[width=0.75\textwidth]{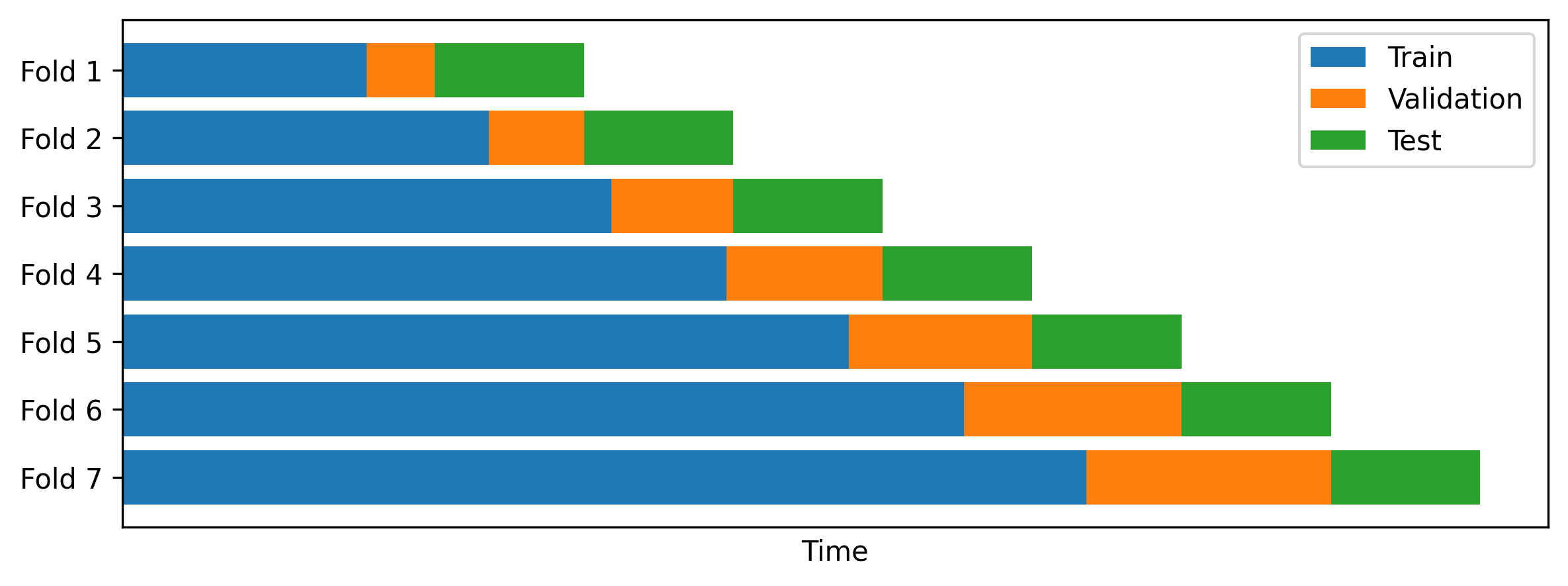}}
	\caption{The implemented time series cross-validation with 7 folds, each consisting of a training, validation, and test period.}
	\label{fig:time_cv}
\end{figure}

A schematic overview of the cross-validation procedure is shown in Figure \ref{fig:time_cv}.
Unlike a fixed-length sliding window, our design follows an expanding-window protocol in which each fold's training set grows monotonically, always incorporating the full historical record available up to the start of that fold.
This approach respects causal ordering, reduces estimation noise for later folds, and reflects the operational regime in which an RL policy deployed in markets accumulates information.

Within each fold, the data are partitioned into training, validation, and test subsets.
The network parameters are updated exclusively on the training portion, hyperparameter tuning and early stopping rely on the validation portion, and final performance metrics are computed on the strictly unseen test portion.

We report the mean and standard deviation of the Sharpe ratio across cross-validation folds to capture not only average performance but also the stability of the learned policies across varying market conditions.

\subsection{Baselines and Comparisons}

The QRL agent is benchmarked against three baselines.
First, the equal-weighted portfolio, which assigns $w_i = \frac{1}{N}$ for all assets and represents a widely used naïve benchmark in portfolio construction for its robustness due to the lack of parameters.
Second, \ac{MVO}, which selects weights that maximize the in-sample Sharpe ratio by brute-force optimization over the admissible allocation space.
\ac{MVO} is theoretically grounded in classical portfolio theory and provides a well-established static standard against which any dynamic strategy should demonstrate value added.

As a comparison, we include classical Deep RL methods: \ac{DDPG} and a continuous-action version of Deep Q-Learning / \ac{DQN} as described in Algorithm~\ref{alg:classical}.
These serve as representatives of current state-of-the-art RL performance.
Importantly, both architectures are evaluated under multiple model sizes to expose the relationship between parameter count, expressivity, runtime, and empirical performance.
Including low-capacity (13.5k parameters), medium-capacity (27k parameters), and high-capacity (160k parameters) classical RL alongside QRL models with 30 and 60 parameters allows us to contrast classical and quantum scaling behaviour.

\subsection{Training and Deployment Setup}

All the classical models are trained on an Nvidia RTX 8000 GPU.
For the quantum models, we adopt a hybrid pipeline in which all training is conducted on a noiseless statevector simulator, and the resulting circuits are deployed to a \ac{QPU} only for inference.
After determining the optimal parameters, the optimized parameterized unitary $\mathbf{U}(\bm{\theta}^\star)$ is executed on a superconducting quantum device to evaluate practical inference behaviour.
Specifically, we use the IBM Eagle r3 \ac{QPU} (127 qubits, 180K CLOPS) to run the converged variational circuit and obtain policy actions via repeated measurement of observable $\hat{B}$ with 10,000 shots.
This deployment step enables us to measure real hardware latency, characterize system-level overhead such as qubit initialization and queueing, and assess the extent to which current quantum devices can support online decision-making in a trading context.

This division between training and execution is motivated primarily by the high cost and limited throughput of current cloud-based QPU access.
At 12 seconds per forward pass and 96 USD per minute IBM's QPU access makes executing the entire training and hyperparameter optimization prohibitively expensive and slow.
A notable downside of this hybrid approach is that statevector simulation scales exponentially with the number of qubits, as opposed to QPU which is practically constant.
This restricts the size of the quantum models that can be feasibly trained.
Our available GPU imposes a practical upper bound on circuit width of roughly 20 qubits, thereby limiting us to somewhat compact \acp{VQC} compared to the 127 qubits that could, in principle, be executed on an IBM Eagle r3.
However, training on a statevector simulator provides the benefit that gradients and expectations can be computed exactly, without the sampling noise that would otherwise arise from finite-shot measurement on hardware. 
As a result, optimization becomes more stable, and convergence is not affected by random variations resulting from the measurement process. 

Gradient-based learning is performed using the parameter-shift rule (see Equation~\ref{eq:param_shift}) in combination with the Adam optimizer \citep{kingma_adam_2017}.
Hyperparameters are tuned using Bayesian search, with the Sharpe ratio on the validation set defined as the acquisition function.
This ensures that the final quantum policies directly target risk-adjusted portfolio performance.
Hyperparameters considered for tuning include learning rates, risk and future reward discount factors, and regularization strength.
Detailed lists of the hyperparameters and their tuning ranges are provided in Appendices~\ref{app:fixed} and \ref{app:tuned}.


\section{Results}

This section examines the results of our empirical evaluation of how quantum agents perform in realistic financial environments and compares their behaviour to established static baselines as well as classical Deep \ac{RL} methods.
Our analysis focuses not only on risk-adjusted performance, but also on robustness across market regimes and efficiency relative to model complexity.

\begin{figure}[H]
    \centering
    \includegraphics[width=0.86\textwidth]{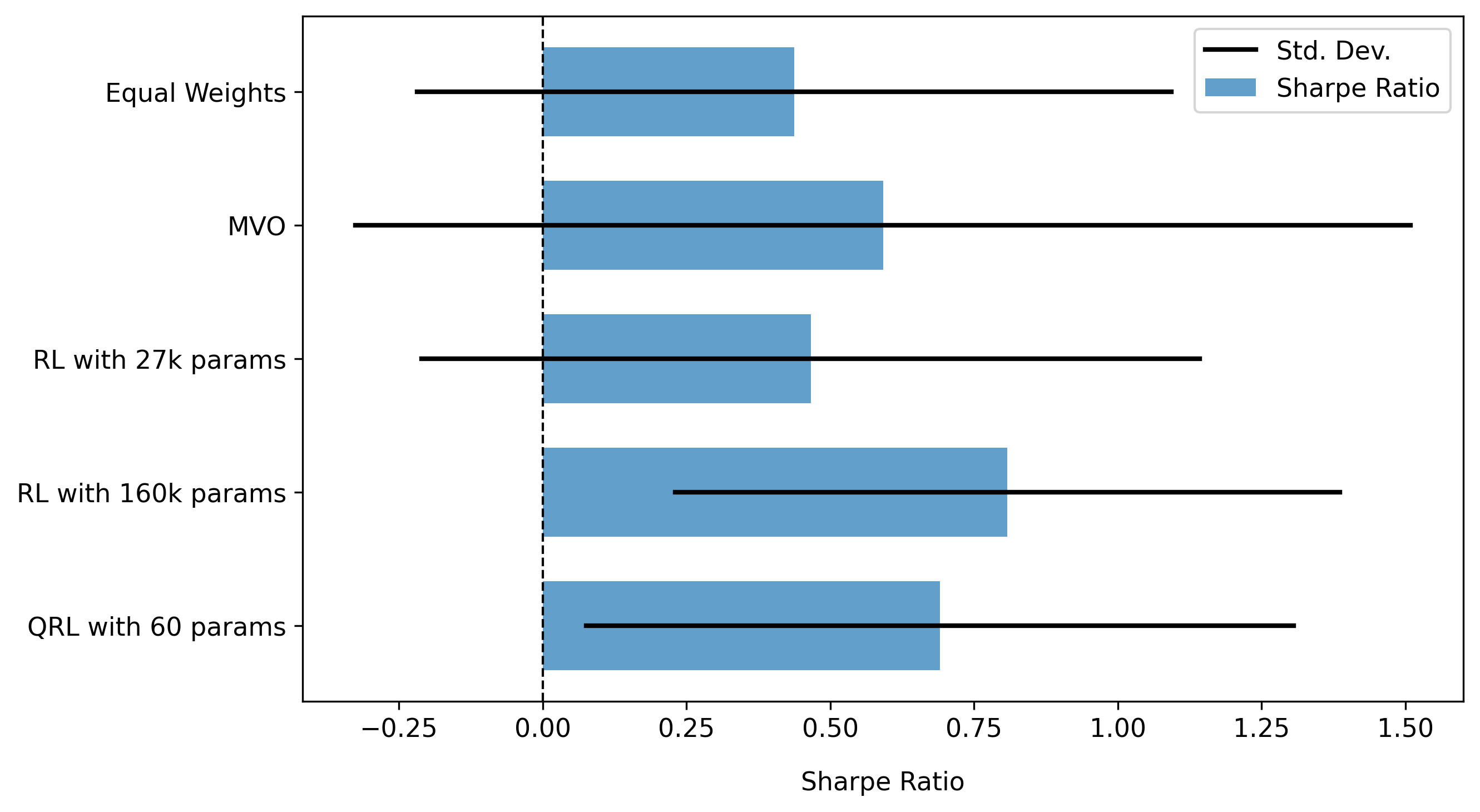}\hspace{15pt}
    \caption{Comparison of portfolio Sharpe ratios across models with error bars indicating variability across cross-validation folds.}
    \label{fig:results_barplot}
\end{figure}

Figure~\ref{fig:results_barplot} summarizes the Sharpe ratios achieved by all models, including static baselines (equal weights and MVO portfolios), classical Deep RL agents with varying parameter counts, and the proposed QRL agent.
Quantum agents consistently outperform classical counterparts when performance is considered relative to parameter count.
It is important to emphasize that the error bars reflect variability across cross-validation folds and therefore measure robustness and stability across market regimes, not statistical uncertainty in the Sharpe ratio estimate itself.

Looking at the disaggregated results provided in Table~\ref{tab:results_table}, the classical models establish a clear scaling trend.
The absolute minimum parameter count in our setup is 13,500 parameters, corresponding to neural networks without hidden layers.
These minimal-capacity agents perform roughly on par with the equal-weights baseline, confirming that this very limited representational flexibility restricts policy quality.
Doubling the model size to 27,000 parameters produces a marginal improvement, whereas the 160k-parameter models reflect the state-of-the-art capacity for classical Deep RL in this environment, achieving the highest Sharpe ratios among all architectures.
Nonetheless, their performance gains come at a steep parameter cost, confirming the inefficiency of classical RL when forced to learn complex stochastic policies.

\begin{table}[H]
	\centering\footnotesize
    \caption{Disaggregated Portfolio Sharpe Ratios of all Models}
	\vspace{0.3em}
	\begin{tabular}{lrr}
		\toprule
		& \hspace{2em}\textbf{Sharpe Ratio\footnotemark[1]} & \hspace{2em}\textbf{~\,Execution Time}\\
		\midrule
		Equal Weights Portfolio & 0.4375 (0.6558) & < 1 s\footnotemark[2]\\
		Mean-Variance Optimization & 0.5919 (0.9170) & < 1 s\footnotemark[2]\\
		\midrule
		Classical DDPG with 13.5k params & 0.4287 (0.6151) & 7 s\footnotemark[2]\\
		Classical DQN with 13.5k params & 0.4446 (0.6568) & 11 s\footnotemark[2]\\[0.5ex]
		\hdashline[1pt/3pt]\noalign{\vskip 0.5ex}
		Classical DDPG with 27k params & 0.4384 (0.6905) & 10 s\footnotemark[2]\\
		Classical DQN with 27k params & 0.4939 (0.6623) & 18 s\footnotemark[2]\\[0.5ex]
		\hdashline[1pt/3pt]\noalign{\vskip 0.5ex}
		Classical DDPG with 160k params & 0.7926 (0.6276) & 14 s\footnotemark[2]\\
		Classical DQN with 160k params\quad & 0.8237 (0.5367) & 27 s\footnotemark[2]\\
		\midrule
		Quantum DDPG with 30 params & 0.4179 (0.6113) & < 1 s (23 min)\textsuperscript{3,4}\\
		Quantum DQN with 30 params & 0.4776 (0.6650) & < 1 s (43 min)\textsuperscript{3,4}\\[0.5ex]
		\hdashline[1pt/3pt]\noalign{\vskip 0.5ex}
		Quantum DDPG with 60 params & 0.7281 (0.7395) & < 1 s (23 min)\textsuperscript{3,4}\\
		Quantum DQN with 60 params & 0.6537 (0.6155) & < 1 s (43 min)\textsuperscript{3,4}\\
		\bottomrule
	\end{tabular}
	\vspace{0.3em}
    \\{\textsuperscript{1}Value in parentheses indicates standard deviation of Sharpe ratio across cross-validation folds.}
	\\[2pt]{\textsuperscript{2}Hardware: Nvidia RTX 8000 GPU.\quad\textsuperscript{3}Hardware: IBM Eagle r3 QPU.}
	\\[2pt]{\textsuperscript{4}Times in parentheses indicate total job duration on cloud-based system.}
    \label{tab:results_table}
\end{table}

Quantum models display a different pattern.
The absolute minimum quantum parameter count is 30, primarily made possible by amplitude encoding, which compresses high-dimensional portfolio states into a compact quantum wavefunction representation.
This compression dramatically reduces qubit requirements while maintaining expressive capacity.
While the 30-parameter quantum agents only barely outperform the equal weights baselines, the advantage of quantum representation emerges when the parameter count is increased.
Doubling the parameters to 60 brings quantum performance into the same range as classical agents with several orders of magnitude more parameters and consistently outperforms MVO.

Although current hardware constraints limited our experiments to quantum models with at most 60 trainable parameters, these compact architectures already achieve performance comparable to, and in some cases exceeding, that of substantially larger classical models.
This result highlights a key advantage of quantum representations in the low-parameter regime: through amplitude encoding and entanglement, \acp{VQC} can capture complex dependencies in high-dimensional financial data without requiring large parameter counts.
Taken together, these results indicate that compact \ac{QRL} models can deliver robust risk-adjusted performance in a low-parameter regime, suggesting that it may offer meaningful advantages even before large-scale, fault-tolerant quantum hardware becomes available.

When looking at the disaggregated results, we observe that \acp{DQN} generally take longer to execute than DDPG due to the overhead associated with sampling multiple candidate actions to approximate the maximum next action-value in continuous action spaces.
Quantum models show moderate variance despite their small parameter footprint, suggesting efficient utilization of representational capacity and a favourable expressivity-to-parameter ratio.

Execution times illustrate the largest operational gap between classical and quantum systems.
Classical agents remain extremely fast in inference, with latencies on the order of seconds when executed on GPU hardware.
Quantum execution, on the other hand, has two faces: The actual evaluation of a VQC (circuit application, unitary evolution, and measurement) is extremely fast, typically on the order of tens of milliseconds, which is why we report an effective inference time of less than one second for all our quantum models when measured strictly at the circuit-execution level.

The substantial delays we observe in end-to-end cloud inference stem not from the intrinsic speed of superconducting quantum hardware, but from the realities of current quantum cloud computing.
On systems such as the IBM Eagle r3, each inference call incurs roughly 12 seconds of latency caused by multi-tenant queueing, repeated qubit initialization, and system-level synchronization steps that must be re-executed at every forward pass because many unrelated users submit heterogeneous circuits to the same device.
These overheads accumulate to total job durations of 23 minutes for quantum DDPG and 43 minutes for quantum DQNs, despite the underlying circuits being rapidly executable.

This divergence between intrinsic quantum computation time and practical end-to-end latency therefore reflects less a fundamental limitation of present-day QPU and more a limitation imposed by the economic and infrastructural constraints of current quantum deployments.
If a QPU were operated in a dedicated, non-cloud-shared setting---where the exact same circuit structure is executed repeatedly, as would be typical for RL inference---the need for repeated calibration, queuing, and reinitialization would be dramatically reduced. Under such conditions, VQC evaluation would approach its native millisecond timescale.

Despite the latency associated with cloud access, the qualitative advantage of quantum models remains clear. Their high expressivity at extremely small parameter counts, combined with favourable asymptotic scaling that avoids the parameter blow-up characteristic of deep classical models, points toward a compelling future trajectory. Classical inference is much faster under current deployment conditions, but as quantum devices become more affordable and accessible (enabling private or semi-dedicated operation with minimal overhead), or cloud-based latencies are reduced, the efficiency frontier may shift.

\section{Conclusion}

Prior research has explored quantum methods for portfolio optimization largely in static settings through QUBO-based formulations, while studies in \ac{QRL} have focused on stylized low-dimensional control problems.
Despite its theoretical potential, the application of \ac{QRL} to complex real-world problems, such as portfolio management, has remained largely unexplored.
This work addresses that gap by developing and examining various \ac{QRL} approaches for dynamic portfolio optimization.
By introducing and benchmarking a fully quantum RL framework for sequential decision-making based on \acp{VQC}, we provide a comprehensive assessment of how quantum agents behave in a meaningful financial setting.

Our QRL agents effectively learn adaptive trading policies that respond to market dynamics.
The presented results show that quantum DDPG and quantum DQNs achieve risk-adjusted performance comparable to, and in some cases exceeding, that of their classical counterparts with several orders of magnitude more parameters.
This efficiency arises from the structural advantages of quantum representation, which enables compact yet expressive policy and value function approximators.

At the same time, our experiments highlight the realities of quantum computing today. While circuit execution on superconducting hardware is extremely fast, cloud-based quantum access introduces substantial latency because many users submit heterogeneous jobs to the same device.
These factors inflate end-to-end inference times from miliseconds to minutes even for shallow circuits.
Importantly, this is not a hardware limitation in the strict sense: if QPUs were operated in dedicated environments, where repeated execution of similar circuits eliminates redundant preparation overhead, inference would approach its native millisecond timescale.

Taken together, these findings place us at an instructive moment for QRL and Quantum Machine Learning in general.
On the algorithmic side, our results demonstrate that QRL already delivers meaningful performance and that quantum expressivity can be leveraged in sequential financial decision-making.
On the hardware side, however, practical deployment is constrained by access models and economic factors.
As quantum devices become more affordable, more stable, and more readily available for dedicated use, the gap between theoretical and practical quantum capability is likely to shrink.
QRL therefore has the potential to transition from a theoretically appealing alternative to a practically viable solution for complex dynamic decision problems in finance and beyond.

\section*{Acknowledgements}

Stefan Lessmann acknowledges financial support through the project ``AI4EFin AI for Energy Finance'', contract number CF162/15.11.2022, financed under Romania's National Recovery and Resilience Plan, Apel nr. PNRR-III-C9-2022-I8.

\clearpage


\bibliographystyle{unsrtnat}
\bibliography{QRLPO}
\urlstyle{same} 


\appendix
\section*{Appendix}
\addcontentsline{toc}{section}{Appendix} 

\renewcommand{\thesubsection}{A.\arabic{subsection}}

\subsection{Fixed Hyperparameters}
\label{app:fixed}

\begin{itemize}
    \item Lookback window: 30 days
    \item Forecast window: 7 days
    \item Soft-update with coefficient \( \tau = 0.005 \)
    \item Epochs: 50, with early stopping enabled at patience 10
    \item Number of DQN next action samples: 10
    \item Replay buffer size: unlimited
    \item Number of shots per QPU circuit evaluation: 10,000
\end{itemize}

\subsection{Hyperparameter Tuning Ranges}
\label{app:tuned}

\begin{itemize}
    \item Actor and critic learning rates \( \in [10^{-4}, 10^{-1}] \), log scale
    \item $\lambda$ (L2 regularization parameter) \( \in [10^{-6}, 10^{-1}] \), log scale
    \item Risk preference \( \in [-1, -10^{-2}] \), linear scale
    \item $\gamma$ (discount factor for future rewards) \( \in [10^{-3}, 10^{-1}] \), log scale
    \item Optimizer \( \in \{\text{Adam}, \text{SGD}\} \)
\end{itemize}

\subsection{Full List of Assets}
\label{app:assets}

\begin{itemize}
    \item AAPL (Apple Inc.)
    \item EFA (iShares MSCI EAFE ETF)
    \item GLD (SPDR Gold Shares)
    \item IWM (iShares Russell 2000 ETF)
    \item JNJ (Johnson \& Johnson)
    \item JPM (JPMorgan Chase \& Co.)
    \item LQD (iShares iBoxx \$ Investment Grade Corporate Bond ETF)
    \item MSFT (Microsoft Corporation)
    \item QQQ (Invesco QQQ Trust)
    \item SPY (SPDR S\&P 500 ETF)
    \item TLT (iShares 20+ Year Treasury Bond ETF)
    \item USO (United States Oil Fund, LP)
    \item XLF (State Street Financial Select Sector SPDR ETF)
    \item XLV (State Street Health Care Select Sector SPDR ETF)
    \item XOM (Exxon Mobil Corporation)
\end{itemize}


\end{document}